\documentclass[10pt,twocolumn,letterpaper]{article}

\usepackage{cvpr}
\usepackage{times}
\usepackage{epsfig}
\usepackage{graphicx}
\usepackage{amsmath}
\usepackage{amssymb}

\usepackage{multirow}
\usepackage{url}
\usepackage{bm}
\usepackage{color}
\usepackage{booktabs}
\usepackage{caption}
\usepackage{authblk}
\usepackage{hyperref}
\hypersetup{
    colorlinks=true,
    linkcolor=blue,
    filecolor=magenta,      
    urlcolor=magenta}

\cvprfinalcopy 

\def\cvprPaperID{****} 

\ifcvprfinal\fi

\begin{document}
\title{ Learning Parallel Dense Correspondence from Spatio-Temporal Descriptors \\
for Efficient and Robust 4D Reconstruction}

\author[1,4]{Jiapeng Tang} 
\author[2]{Dan Xu}  
\author[1,5,6]{Kui Jia \thanks{Corresponding author}}
\author[3,4]{Lei Zhang}
\affil[1]{School of Electronic and Information Engineering, South China University of Technology}
\affil[2]{Department of Computer Science and Engineering, HKUST, HK}
\affil[3]{Department of Computing, The Hong Kong Polytechnic University, HK}
\affil[4]{DAMO Academy, Alibaba Group}
\affil[5]{Pazhou Lab, Guangzhou, China} 
\affil[6]{Peng Cheng Laboratory, Shenzhen, China}
\affil[ ]{{\tt\small msjptang@mail.scut.edu.cn, \tt\small danxu@cse.ust.hk, \tt\small kuijia@scut.edu.cn, \tt\small cslzhang@comp.polyu.edu.hk}}

\maketitle
\thispagestyle{empty}

\begin{abstract}
    This paper focuses on the task of 4D shape reconstruction from a sequence of point clouds. Despite the recent success achieved by extending deep implicit representations into 4D space~\cite{niemeyer2019occupancy}, it is still a great challenge in two respects, i.e.~how to design a flexible framework for learning robust spatio-temporal shape representations from 4D point clouds, and develop an efficient mechanism for capturing shape dynamics. In this work, we present a novel pipeline to learn a temporal evolution of the 3D human shape through spatially continuous transformation functions among cross-frame occupancy fields. The key idea is to parallelly establish the dense correspondence between predicted occupancy fields at different time steps via explicitly learning continuous displacement vector fields from robust spatio-temporal shape representations.
    Extensive comparisons against previous state-of-the-arts show the superior accuracy of our approach for 4D human reconstruction in the problems of 4D shape auto-encoding and completion, and a much faster network inference with about 8 times speedup demonstrates the significant efficiency of our approach. The trained models and implementation code are available at {\url{ https://github.com/tangjiapeng/LPDC-Net}}.
\end{abstract} 

\section{Introduction}
\label{SecIntro}
\begin{figure}[t]
    \vspace{-6pt}
    \centering 
        \includegraphics[scale=0.45]{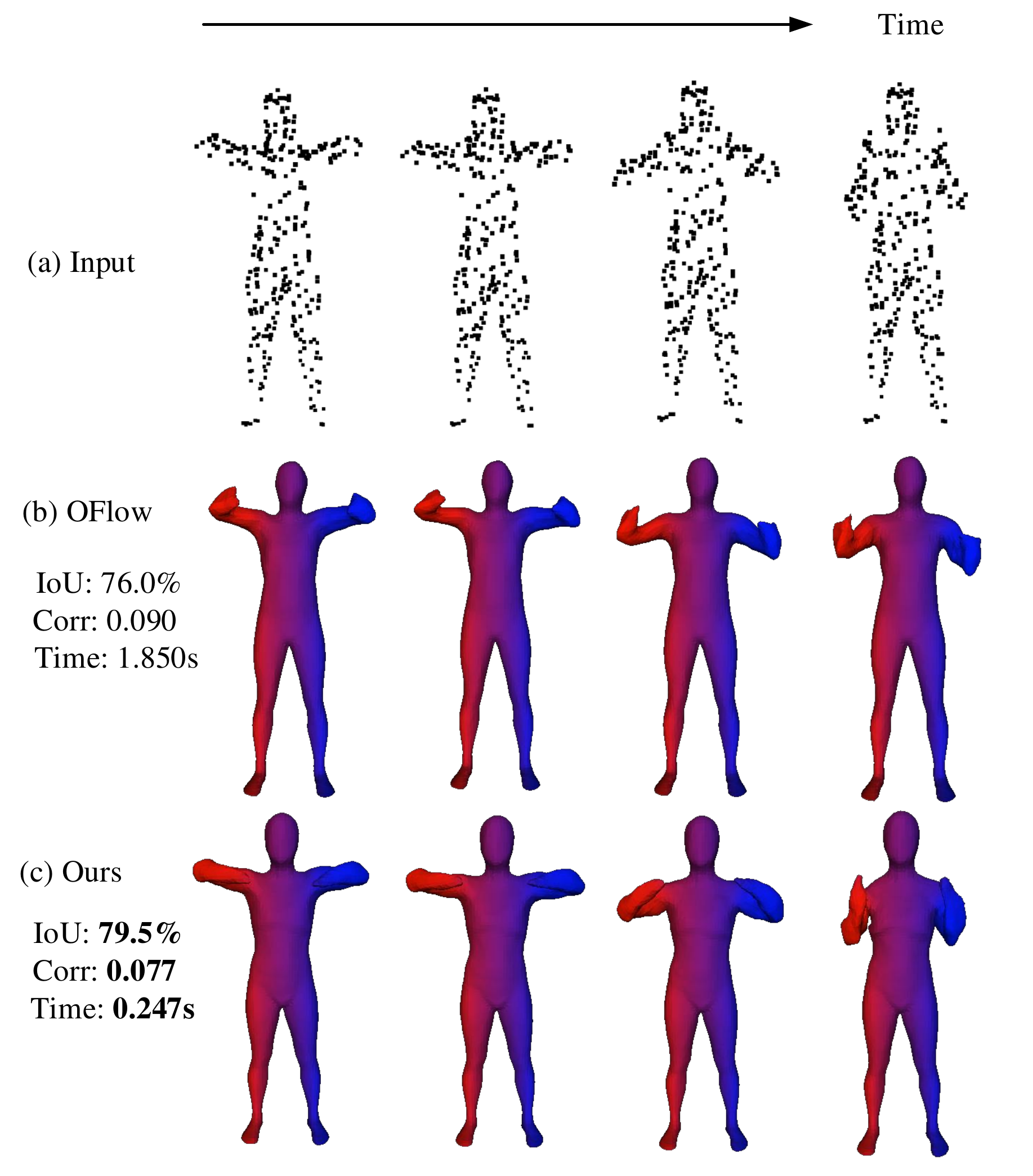}
        \vspace{-15pt}
        \caption{Given a sequence of 3D point clouds sampled in space and time, our goal is to reconstruct time-varying surfaces with dense correspondences. Compared to the state-of-the-art, \ie~OFlow~\cite{niemeyer2019occupancy}, our approach can obtain more accurate geometries (higher IoU), better coherence (lower correspondence error) while supporting $8$x faster inference.} 
    \vspace{-12pt}
    \label{fig:teaser}
\end{figure}

We are surrounded by spatio-temporally changing environments that consist of various dynamics, such as observer movements, object motions, and human articulations. Reconstructing the human bodies evolving over time is vital for various application scenarios such as robot perception, autonomous driving, and virtual/augmented reality. 

Traditional works have achieved varying degrees of success in learning 4D reconstruction (\ie 3D reconstruction along time) from a temporal sequence of point clouds, they are faced with various restrictions including the requirement of an expensive template mesh~\cite{alldieck2018video, huang2017towards, tung2017self, kanazawa2019learning, kocabas2020vibe} or the dependence on smooth and clean inputs in space and time~\cite{wand2007reconstruction}. 
To overcome these issues, OccFlow~\cite{niemeyer2019occupancy} proposes a learning-based 4D reconstruction framework that establishes dense correspondences between occupancy fields by calculating the integral of a motion vector field defined in space and time to implicitly describe the trajectory of a 3D point.  Although impressive results have been achieved, there are still several inherent limitations in this framework. Firstly, its spatial encoder does not take into account the aggregation of shape properties from multiple frames, which degrades the capability to recover accurate surface geometries. In addition, its temporal encoder ignores the time information which is of great importance to capture the temporal dynamics. Secondly, the integral of estimated immediate results leads to accumulated prediction errors in the temporal continuity and the reconstructed geometries. Lastly, it demonstrates low computational efficiency during training and inference because of the demanding computations of solving complex neural ordinary differential equations~\cite{chen2018neural} to sequentially calculate the trajectories of points over time.

To tackle the above-mentioned problems, we aim to design a novel framework for 4D shape reconstruction from spacetime-sampled point clouds, to advance the 4D reconstruction from computational efficiency, accurate geometry, and temporal continuity. Our key idea is a mechanism which \emph{parallelly} establishes the dense correspondence among different time-step occupancy fields predicted from the learned robust spatio-temporal shape representations. A high-level design of our proposed approach is a combination of static implicit field learning and dynamic cross-frame correspondence predicting. 
The former one focuses on occupancy field predictions from a novel spatio-temporal encoder that can effectively aggregate the shape properties with the temporal evolution to improve the robustness of geometry reconstructions. The latter one is utilized to identify the accurate correspondences within cross-frame occupancy fields,  which are produced from representative embeddings describing the spatio-temporal changes in an efficient manner. The key to achieving this goal is a strategy of simultaneously learning occupancy field transformation from a first time step to others. It can help to remarkably reduce the convergence time in the network training, because of the bypassing of the expensive computation caused by solving ordinary differential equations. Moreover, benefiting from the advantages of parallel isosurface deformations for the different time steps, our method provides a significant speed-up of the inference time. As shown in Fig.~\ref{fig:teaser}, we can achieve more robust surface reconstructions and more accurate correspondence prediction while allowing for considerably faster inference.
\par The main contribution can be summarized as follow:
\begin{itemize}
\item{We propose a learning framework of modeling the temporal evolution of the occupancy field for 4D shape reconstruction, which is capable of capturing accurate geometry recoveries and coherent shape dynamics.}
\item{We develop a novel strategy of establishing cross-frame shape correspondences by paralleling modeling occupancy field transformations from the first frame to others, which significantly improves the network computation efficiency, especially in the inference stage.}
\item{We propose a novel 4D point cloud encoder design that performs efficient spatio-temporal shape properties aggregation from 4D point cloud sequences, which improves the robustness of reconstructed geometries.}
\end{itemize}
Extensive ablation studies are conducted to validate the effectiveness of our proposed module designs. Comparisons against previous state-of-the-arts on the challenging D-FAUST dataset demonstrate the superior accuracy and efficiency of our approach in the problems of 4D shape auto-encoding and completion.

\section{Related Work}
    In this section, we review the closely related works from three aspects as follows. 

    \begin{figure*}[t]
        \centering
        \includegraphics[scale=0.60]{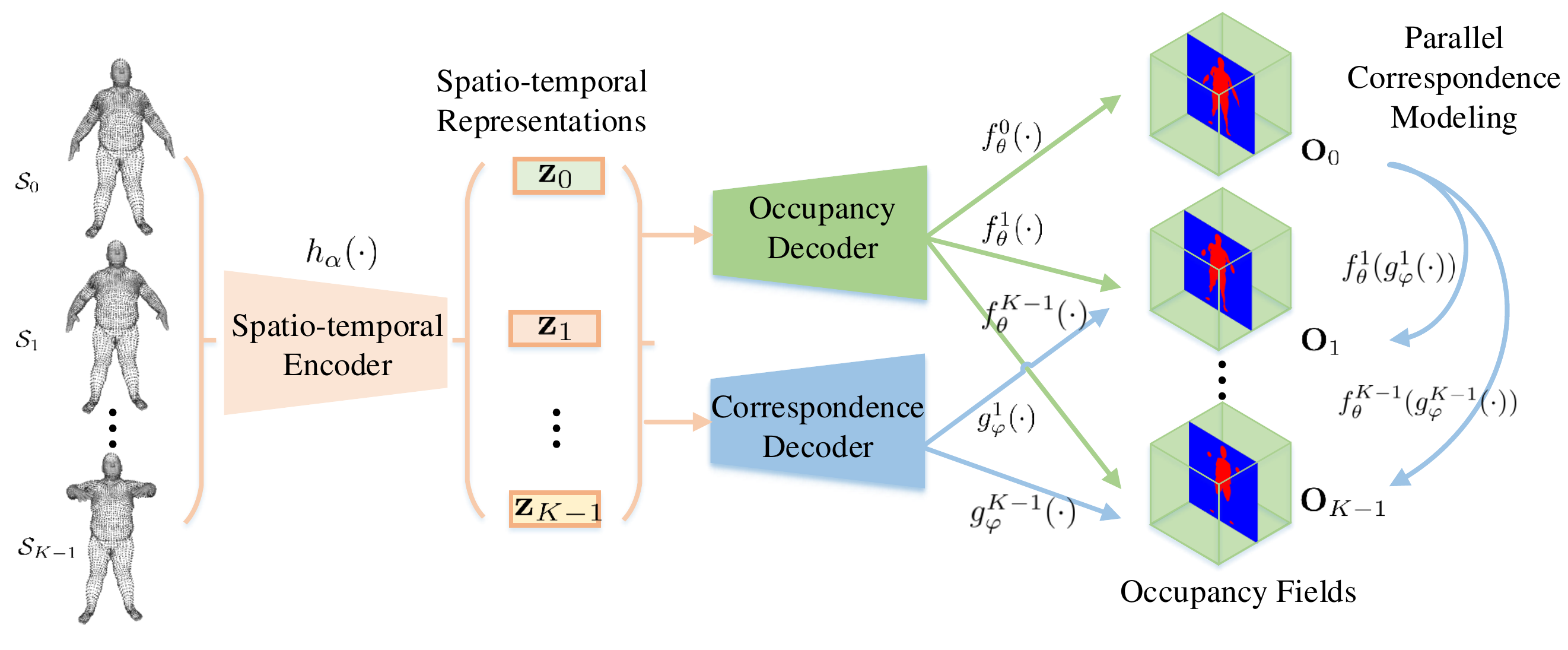}
        \vspace{-5pt}
        \caption{\textbf{Model Overview}. The proposed model first inputs a 3D point cloud sequence $\{ \mathcal{S}_0, \mathcal{S}_1, ..., \mathcal{S}_{K-1}  \}$ into a designed spatio-temporal encoder which extracts latent representations $\{ \mathbf{z}_0, \mathbf{z}_1, ... \mathbf{z}_{K-1} \} $. And then, the representations go through two separate decoders, \ie~the occupancy and the correspondence decoder. The occupancy decoder targets predicting the occupancy fields $\mathbf{O}_0, \mathbf{O}_1, ..., \mathbf{O}_{K-1}$ in each frame. Finally, the correspondence decoder is used to parallelly model the correspondence between $\mathbf{O}_0$ of the first frame and $\mathbf{O}_1, ..., \mathbf{O}_{K-1}$ of others.}
    \label{fig:pipeline}
    \vspace{-10pt}
\end{figure*}
    \noindent \textbf{3D Shape Reconstruction}
    The commonly used shape representations include voxel~\cite{choy20163d, han2017high}, octree~\cite{riegler2017octnet, tatarchenko2017octree, wang2018adaptive, hane2017hierarchical}, point cloud~\cite{fan2017point} , mesh~\cite{groueix2018atlasnet, wang2018pixel2mesh, kato2018neural, pan2019deep, tang2019skeleton}, implicit field~\cite{chen2019learning, mescheder2019occupancy, park2019deepsdf, xu2019disn, peng2020convolutional, yang2020deep, zhao2020sign}, and hybrid representations~\cite{tang2019skeleton, tang2020skeletonnet, poursaeed2020coupling}. 
    Especially, the implicit representations~\cite{chen2019learning, mescheder2019occupancy, park2019deepsdf, xu2019disn, peng2020convolutional}, which implicitly represent a 3D surface by a continuous function, have attracted much attention. The continuity enables more accurate surface recoveries and volumetric reconstruction with infinite resolution. To inherit the advantages, we propose to extend the implicit representation for 4D reconstruction. Instead of independently extracting a surface mesh from the implicit field of each frame in the input sequence, which typically leads to slow inference and non-consistent topologies, we avoid these issues by the dense correspondence modeling which propagates the extracted surface mesh from the initial state to others.

    \noindent\textbf{Dynamic 4D Reconstruction} Traditional works~\cite{leroy2017multi, mustafa2015general, mustafa2016temporally, neumann2002spatio, starck2007surface} utilize multi-view geometry to tackle the dynamic scene reconstruction problem from videos captured by multiple cameras. In contrast to them, we aim at 4D shape reconstruction from a sequence of dynamically scanned point clouds, while the existing works with similar settings are faced with various limitations, including a heavy dependence on spatio-temporally smooth inputs~\cite{wand2007reconstruction} or the requirement of expensive template meshes~\cite{alldieck2018video, huang2017towards, tung2017self, kanazawa2019learning, kocabas2020vibe}. 
    Compared to OccFlow~\cite{niemeyer2019occupancy}, instead of predicting the motion vectors for points in space and time and relying on the solver of Neural ODE~\cite{chen2018neural} to calculate their 3D trajectories, we {directly} model the movements of points, which decreases the computation overhead during training. And the supporting of parallel surface deformation at different time steps remarkably accelerates the inference speed.
    Besides, we design a unified spatio-temporal encoder to effectively capture temporal dynamics and utilize the important time information in learning spatio-temporal descriptors.
    
    \noindent \textbf{Shape Correspondence Modeling}
    Modeling point-to-point correspondence between two 3D shapes~\cite{biasotti2016recent, van2011survey, tam2012registration} is a well-studied area in computer vision and graphics. 
    Our goal of modeling time-varying occupancy fields is closely related to deformation field-based methods~\cite{luthi2017gaussian, myronenko2010point}. However, most of these works only define vector fields on the surfaces rather than in the whole 3D space as us.  Eisenberger~\etal~\cite{slavcheva2017towards} choose to model the evolution of the signed distance field to implicitly yield correspondences. They optimize an energy function in the evolution equation to impose similarity relationships of the Laplacian eigenfunction representations between the input and target shapes. However, we learn the dense correspondences between time-varying occupancy fields based on an intuitive observation, namely that the occupancy values of points are always invariant along the temporal evolution.
\label{SecRelated}
\section{Approach}
\label{SecApproa}
\begin{figure*}[h]
    \centering
        \includegraphics[scale=0.6]{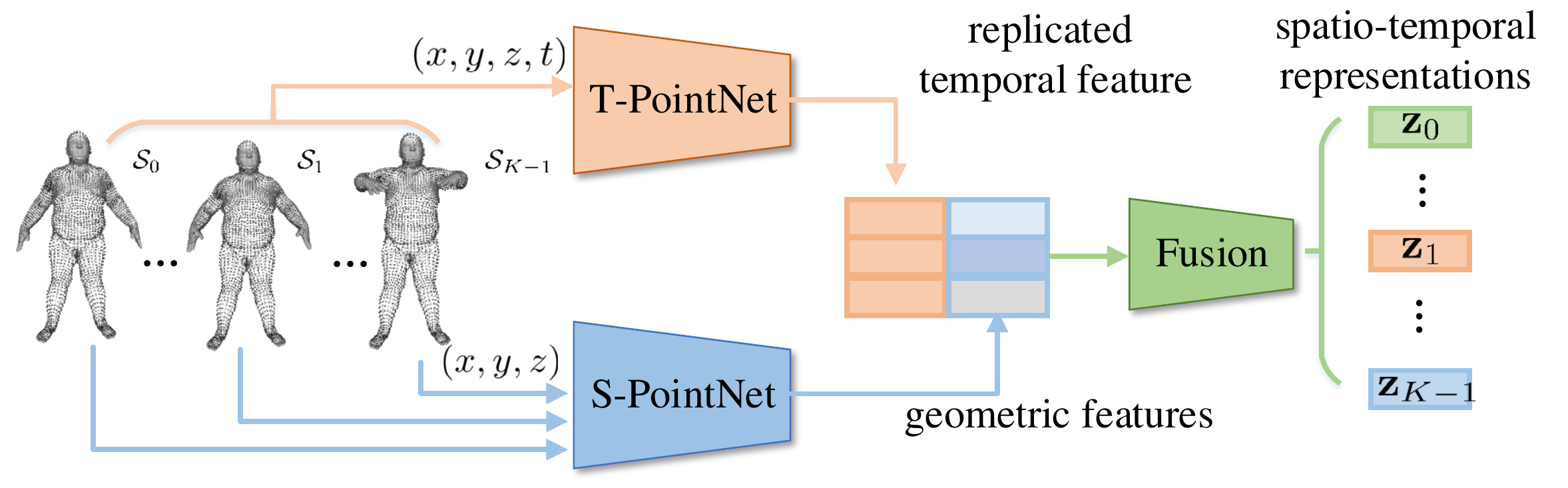}
        \vspace{-5pt}
        \caption{\textbf{Spatio-Temporal Encoder}. It contains a T-PointNet branch that utilizes the entire 4D point cloud $\mathcal{S} = \{ {\mathcal{S}_k} \}_{k=0}^{K-1}$ to extract a temporal representation by treating time explicitly and equally to each spatial dimension. It also uses a S-PointNet branch to extract a sequence of geometric representations by individually applying it for each frame $\mathcal{S}_k$ without considering timestamps. Finally, the geometric and temporal representations are fused into a sequence of spatio-temporal descriptors $\{ \mathbf{z}_0, \mathbf{z}_1, ... \mathbf{z}_{K-1} \} $, to aggregate shape properties and explore dynamics variations.}
        \vspace{-5pt}
    \label{fig:encoder}
    \vspace{-4pt}
\end{figure*}

In this section, we first formulate the dynamic 4D surface reconstruction problem as the following. We consider as input a sequence of potentially incomplete, noisy 3D point clouds ( of human bodies, easily captured by depth sensors). The observation of each frame can be represented as a point set $\mathcal{P} = \{ \mathbf{p}_i \in \mathbb{R}^3 = \{ x_i, y_i, z_i \} | i = 0, 1, ... \}$. For a sequence of $K$ frames with possibly non-uniform time intervals, we consider it as a 4D spatio-temporal point cloud donated by $\mathcal{S} = \{ {\mathcal{S}_k} \}_{k=0}^{K-1}$, where ${ \mathcal{S}_k } = \{ \mathbf{s}_i \in \mathbb{R}^4 = \{ x_i, y_i, z_i, t_k \}_{i=0}^{{N_k}-1} \}$. ${N_k}$ is the number of points at frame $k \in [0, K-1]$ and at time $t_k \in [{t_0}, {t_{K-1}}] \subset \mathbb{R}$ with ${N} = \sum_{k=1}^K N_k$. Our goal is to reconstruct time-varying 3D surfaces with accurate geometry, temporal coherence and fast inference. We achieve this by (1) developing our model based on the implicit surface representation which has been demonstrated the impressive capacity of capturing complex object geometries \cite{ park2019deepsdf, xu2019disn, atzmon2020sal, xu2019geometry, gropp2020implicit} (2) capturing the robust spatio-temporal shape properties by efficiently fusing information from each frame and taking the time information into consideration to obtain accurate temporal dynamics (3) parallelly modeling the dense correspondences between cross-time occupancy fields, which facilities parallel surface deformations from the first to other time steps to accelerate the inference speed.

\noindent \textbf{Overview}  The overall pipeline is shown in Fig. \ref{fig:pipeline}. It is composed of three key components which are respectively responsible for spatio-temporal representation learning, occupancy fields predicting, and dense correspondences modeling.  We firstly process the input (\ie~ a point cloud sequence $\mathcal{S}$) through a spatio-temporal encoder $h_\alpha(\cdot)$ to get a sequence of latent embeddings $\{ \mathbf{z}_0, \mathbf{z}_1, ... \mathbf{z}_{K-1} \} $ encoding the geometric shape properties and the temporal changes. Then we learn the occupancy field $f_\theta^{k}(\cdot)$ at $t=t_k$ using a shared decoder $f_\theta : \mathbb{R}^3 \rightarrow [0, 1]$ conditioned on a time-specific latent embedding $ \mathbf{z}_k$. And a correspondence decoder $g_\varphi : \mathbb{R}^3 \rightarrow \mathbb{R}^3$ is utilized to  simultaneously estimate continuous displacement vector fields to model occupancy field evolutions from initial to future time steps. More specifically, it establishes dense correspondences between $f_\theta^{0}(\cdot)$ and $f_\theta^{k}(\cdot)$ by learning a function $g_\varphi^k(\cdot)$ that transforms 3D spatial points at time $t_0$ into coordinate system at time $t_k$  conditioned on the associate embeddings $\mathbf{z}_0$ and $ \mathbf{z}_k$. 
So the occupancy field at time $t_k$ can be predicted through $f_\theta^k (g_\varphi^k(\cdot))$. The $\alpha$, $\theta$ and $\varphi$ respectively denote learnable network parameters of $h(\cdot)$, $f^k(\cdot)$ and $g^k(\cdot)$. In the following, we explain more details about the spatio-temporal encoder (Section \ref{SubSecTemp}), the occupancy field decoder (Section \ref{SubSecOcc}), the correspondence decoder (Section \ref{SubSecCorr}), the training paradigm (Section \ref{SubSecTrain}), and the inference (Section \ref{SubSecInfer}).

\subsection{Spatio-temporal Representations Learning}
\label{SubSecTemp}
To understand 3D shape motions in a sequence of consecutive observed frames, it is crucial to learn both geometric features for shape recovery and the temporal correlation for continuity maintenance. A straightforward solution is to follow the previous method that~\cite{niemeyer2019occupancy} uses two parallel PointNet-based ~\cite{charles2017pointnet}  encoders to extract shape and motion embeddings respectively. However, its proposed shape encoder only uses the first point cloud $\mathcal{S}_0$ to acquire the shape feature. Thus the reconstructed surfaces are always sub-optimal if $\mathcal{S}_0$ is seriously incomplete, as the shape properties in other frames can not be incorporated.  And its proposed temporal encoder strictly assumes that the point-wise correspondence between different input frames is known in the learning, which restricts its flexibility of processing real scans without this relationship. Moreover, the aggregation of temporal information does not explicitly take into account the time information.  Our motivation is to aggregate the shape features from different frames to capture robust embeddings for the implicit surface generation, and to treat time as important as the spatial coordinates to capture expressive embeddings for describing the dynamic shape evolution.  Although it is possible to utilize the technique of 4D point cloud processing proposed by~\cite{liu2019meteornet}, it would be insufficient due to time-consuming spatio-temporal neighborhood queries. Thus we introduce a novel spatio-temporal encoder shown in Fig.~\ref{fig:encoder}. Specifically, it contains a T-PointNet branch that accepts and transforms the entire 4D point cloud $\mathcal{S} = \{ {\mathcal{S}_k} \}_{k=0}^{K-1}$ to extract a temporal representation by treating time explicitly and equally to each spatial dimension. It also uses the S-PointNet branch to produce a sequence of geometric representations by individually applying it for each frame $\mathcal{S}_k$ without considering timestamps. Finally, the geometric and temporal features are fused to obtain a sequence of descriptors $\{ \mathbf{z}_0, \mathbf{z}_1, ... \mathbf{z}_{K-1} \}$ aggregating shape properties and exploring dynamics variations. 

\subsection{Occupancy Fields Predicting}
\label{SubSecOcc}
  In this section, we present the details of the learning occupancy field at each time step for 4D shape reconstruction. 
  The occupancy field represents the 3D shape as a continuous boundary classifier where each 3D point is classified as 0 or 1, depending on whether the point lies inside or outside the surface.
  Although the signed distance field (SDF) can be an alternative choice, we observed convergence issues by employing encoder-decoder architecture for SDF learning from sparse point clouds. 
  
  According to the universal approximation theorem~\cite{hornik1989multilayer}, we implement the occupancy field learning as a multi-layer perceptron (MLP) to predict occupancy states for the points sampled in space and time. Given a point $\mathbf{p}_k$ sampled at time $t_k$, the probability of locating outside the 3D human body is predicted by $f_\theta^k(\mathbf{p}_k) := f(\mathbf{p}_k; \mathbf{z}_k)$ that feed-forwards the feature constructed by concatenating the point coordinates $\mathbf{p}_k$ and its associate spatio-temporal feature $\mathbf{z}_k$. 

\begin{figure}[h]
    \vspace{-6pt}
    \centering
        \includegraphics[scale=0.7]{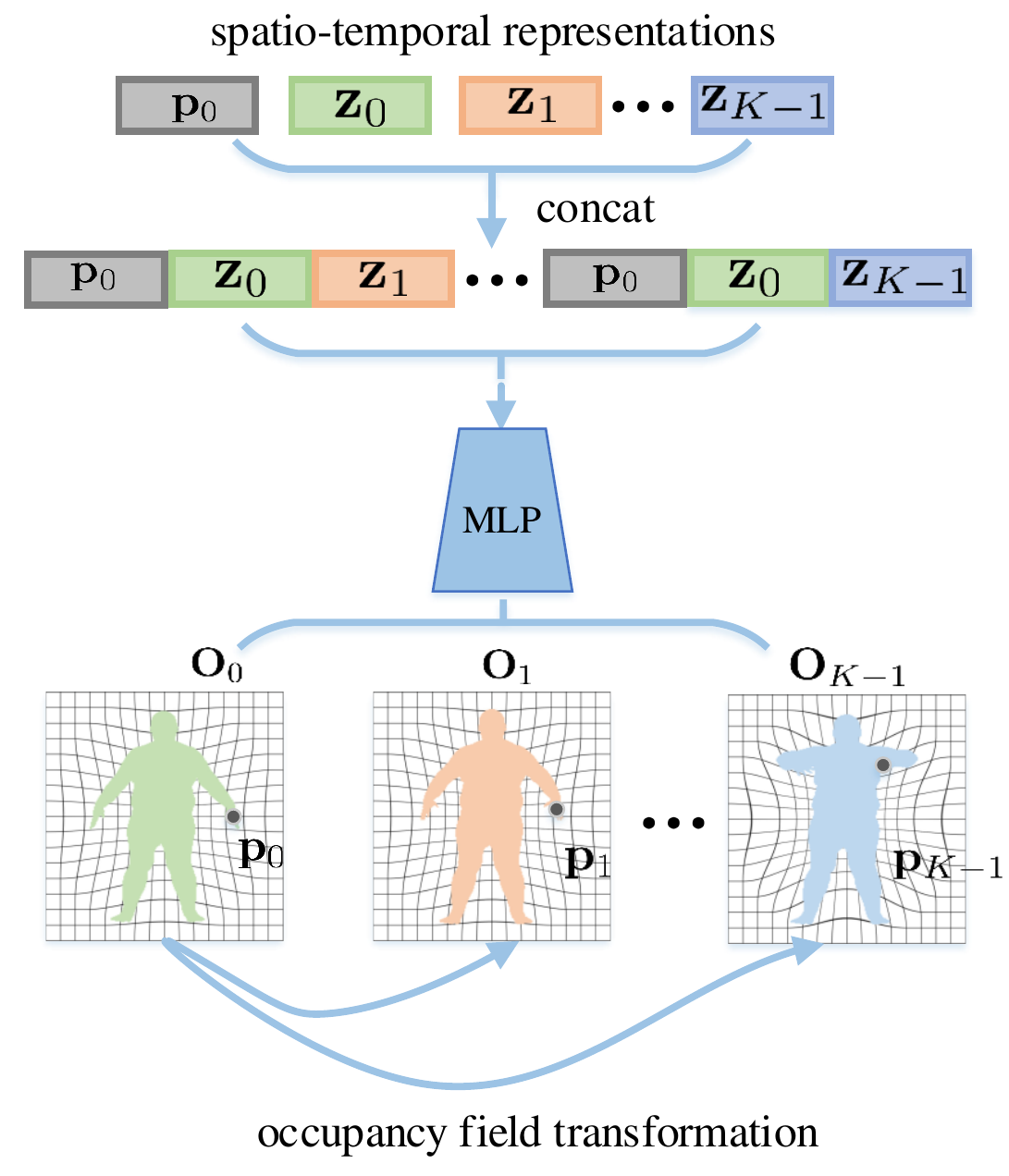}
        \vspace{-0.2cm}
        \caption{\textbf{Dense Correspondence Decoder.}
         Based on the learned spatio-temporal representations $\{ \mathbf{z}_0,\mathbf{z}_1, ...,\mathbf{z}_{K-1} \}$, the correspondence decoder utilizes a shared MLP conditioned on the concatenation of the associate representations  $\mathbf{z}_0$ and $\mathbf{z}_k$, to predict the displacements from $\mathbf{p}_0$ at time $t_0$ to $\mathbf{p}_k$ located in the coordinate system at time $t_k$}.
    \label{fig:decoder}
    \vspace{-14pt}
\end{figure}

\subsection{Cross-time Correspondence Modeling}
\label{SubSecCorr}
In this section, we describe the details about dense correspondence modeling between the initial occupancy field $\mathbf{O}_0$ and others ( $\mathbf{O}_1, ..., \mathbf{O}_{K-1}$ ).
Although it is feasible to model the occupancy field transformation between consecutive two frames in order to omit the complex computations of solving neural ODE ~\cite{chen2018neural}, the sequential manner would lead to accumulated prediction errors and slow inference. Thus we choose to implement this by predicting displacement vector fields to future time steps in parallel paths.

Each displacement vector function is responsible for describing the occupancy field deformation from initial frame to subsequent frames. 
More specifically, the transformation process from time $t_0$ to time $t_k$ can be formulated as:
\begin{eqnarray}
      {\mathbf{p}_k} = {\mathbf{p}_0} + g_\varphi^k({\mathbf{p}_0})
\end{eqnarray}
where $g_\varphi^k: \mathbb{R}^3 \rightarrow \mathbb{R}^3$ is used to predict the displacement of each point $\mathbf{p}_0$ at time $t_0$ to the associate position $\mathbf{p}_k$ located in the coordinate system at time $t_k$. According to Motion Coherent Theory~\cite{yuille1989mathematical}, it is significant to ensure the deformation vector function $g_\varphi^k$ are continuous. To meet this goal, we implement $g_\varphi^k$ with a shared multi-layer perceptron (MLP) conditioned on $\mathbf{z}_0$ and $\mathbf{z}_k$  that capture geometric properties and temporal dynamics at time $t_0$ and $t_k$. 
The dense correspondence decoder is shown in Fig.~\ref{fig:decoder}. Specifically, for each point $\mathbf{p}_0$, we concatenate its coordinates with the spatiotemporal features $\mathbf{z}_0, \mathbf{z}_i$ that are associated with time $t_0$ and $t_k$. Then the displacement vector $\mathbf{p}_0$ to $\mathbf{p}_k$ can be obtained by
\begin{eqnarray}
     \mathbf{p}_k - \mathbf{p}_0 = g_\varphi( \mathbf{p}_0 \oplus \mathbf{z}_0 \oplus \mathbf{z}_k)  
\end{eqnarray}
{where the symbol $\oplus$ denotes a concatenation operation along the feature channel direction.}

\subsection{Training Objective}
\label{SubSecTrain}
Our network learning is supervised by two types of optimization losses. For the occupancy field generation and transformation, we employ the standard binary cross-entropy loss for measuring the discrepancy between the predicted probabilities and the ground truths. It is defined as:
\begin{eqnarray}
\label{EqnSDFLoss}
      \mathcal{L}_{occ} = {\sum_k}{\sum_{\mathbf{p}_k \in \mathcal{P}_k}}  \mathcal{L}_{bce}(f_\theta^k(\mathbf{p}_k), \mathbf{O}^k(\mathbf{p}_k)) + \\ \nonumber \quad\quad
      \mathcal{L}_{bce}(f_\theta^k (g_\varphi^k(\mathbf{p}_0)), \mathbf{O}^k(g_\varphi^k(\mathbf{p}_0))),
\end{eqnarray}
where $\mathbf{O}^k(\mathbf{p}_k)$ denotes the ground truth occupancy value of $\mathbf{p}_k$ at time $t_k$.
The first term is used to constrain the implicit surface generation at each time. And the second term is used to constrain the occupancy states changing for non-surface points.

\par The dense correspondence decoder is also trained by constraining the temporal evolution of 3D points sampled from the dynamic surfaces.
The temporal correspondence loss can then be defined as:
\begin{eqnarray}
\label{EqnCorrLoss}
      \mathcal{L}_{corr} =  {\sum_k} {\sum_{ \mathcal{Q}}}  |\bm{g}_\varphi^k(\mathcal{Q}(t_0)) - \mathcal{Q}(t_k)| 
\end{eqnarray}
where $\mathcal{Q}$ denotes the a trajectory $\mathcal{Q}(t_0)$, $\mathcal{Q}(t_1)$, ..., $\mathcal{Q}(t_{K-1})$ sampled from the dynamic surfaces at different time steps. 

Then the overall optimization objective of our proposed approach $\mathcal{L}_{total}$ can be formulated as follows:
\begin{eqnarray}
\label{EqnTotalLoss}
      \mathcal{L}_{total} = {\mathcal{L}}_{occ} + \lambda * {\mathcal{L}}_{corr}
\end{eqnarray}
where $\lambda$ is a hyper-parameter weighting the importance of the temporal correspondence loss $\mathcal{L}_{corr}$.

\subsection{Inference}
\label{SubSecInfer}
During the inference stage, we predict the dynamic 3D shapes for a new observation $\mathcal{S}$ by first reconstructing the shape at starting time $t = t_0$, followed by propagating the reconstruction into the future $t \in [t_1, ... t_{K-1}]$ using the trained correspondence decoder. Thus we do not need to predict the occupancy field at each time step. We use the Multiresolution IsoSurface Extraction (MISE) \cite{mescheder2019occupancy} and marching cubes algorithms~\cite{lorensen1987marching, icml2020} to extract the triangular mesh ${\mathcal{M}_0} = \{ {\mathcal{V}_0}, {\mathcal{E}_0}, {\mathcal{F}_0} \}$ where ${\mathcal{V}_0}, {\mathcal{E}_0}, {\mathcal{F}_0}$ represent the vertices, edges, and faces of mesh $\mathcal{M}_0$ from the predicted occupancy field at initial time $t = t_0$. For other time steps in the future, we use the learned deformation vector fields to calculate the displacement $g_\varphi^k(\mathbf{v}_i)$ for the each vertice $\mathbf{v}_i$ in ${\mathcal{V}_0}$ while fixing the topology connectivity relationships ${\mathcal{E}_0}, {\mathcal{F}_0}$. So the mesh at time $t_k$ is obtained through:
\begin{eqnarray}
      {\mathcal{M}_k} = \{ \{ g_\varphi^k(\mathbf{v}_i) | \mathbf{v}_i \in \mathcal{V}_0  \},  \mathcal{E}_0, \mathcal{F}_0\}
\end{eqnarray}
Notably, compared with previous methods \cite{mescheder2019occupancy, niemeyer2019occupancy}, our approach can provide a faster network inference as it parallelly estimates the vertex displacements of $\mathcal{M}_0$ for different time steps and avoids the expensive computation of solving ordinary differential equations. 
\section{Experiment}
\label{SecExper}
    \noindent \textbf{Datasets}
    Our experiments are performed on the challenging Dynamic FAUST (D-FAUST)~\cite{bogo2017dynamic} dataset which contains raw-scanned and registered meshes for 129 sequences of 10 humans (5 females and 5 males) with various motions such as “shake hips”, “running on spot”, or “one leg jump”. 
    Same as the train/val/test split of OFlow~\cite{niemeyer2019occupancy}, we divide all sequences into training (105), validation (6), and test (21) sequences. 
    All models are evaluated on unseen actions or individuals during training. The test set consists of two subsets. One (S1) contains 9 sequences of seen individuals with unseen motions in the train set. The other (S2) contains 12 sequences of an unseen individual. To increase the size of the training samples, we subdivide each sequence into short segments of 17 time steps or long segments of 50 time steps according to different experiment settings.

    \noindent \textbf{Baselines}
    We compare our approach with three state-of-the-arts for 4D reconstruction from point cloud sequences, including PSGN 4D, ONet 4D, and OFlow. The PSGN 4D extends the PSGN~\cite{fan2017point} to predict a 4D point cloud, \ie~the point cloud trajectory instead of a single point set. The ONet 4D is a natural extension of ONet~\cite{mescheder2019occupancy} to  define the occupancy field in the spatio-temporal domain by predicting occupancy values for points sample in space and time.
    The OFlow~\cite{niemeyer2019occupancy} assigns each 4D point an occupancy value and a motion velocity vector and relies on the differential equation to calculate the trajectory.
    For a fair comparison, we train all models of baselines with the paradigms in~\cite{niemeyer2019occupancy}.

    \noindent \textbf{Implementation details}
    {The model implementation is based on OFlow~\cite{niemeyer2019occupancy}.} 
    For all experiments, our model is trained in an end-to-end manner using a batch size of 16 with a learning rate of $1e-4$ for 400k iterations. 
    For the loss calculation in each training iteration, we randomly sample a fixed number of points in space and time. More specifically, for the occupancy prediction loss $L_{occ}$, we sample 512 points that are uniformly distributed in the bounding box of the 3D shapes at each respective time. For the correspondence loss $L_{corr}$, we uniformly sample the trajectories of 100 points from the sequence of ground truth surfaces. And the hyperparameters used in Equation~\ref{EqnTotalLoss} are $\lambda = 1$.
    
    \noindent \textbf{Evaluation Metrics}
    We use Chamfer distance (lower is better), Intersection over Union  (higher is better), and correspondence distance  (lower is better) as primary metrics to evaluate the reconstructed surface mesh sequences. {We follow OFlow ~\cite{niemeyer2019occupancy} to compute these evaluation metrics.}
    
   \subsection{4D Shape Reconstruction}
            \begin{table}[t]
    \renewcommand\arraystretch{1.2}
    \setlength{\tabcolsep}{2mm}
            \centering
            \begin{tabular}{c | c | c c c}
            \toprule
             & Method &  IoU & Chamfer & Correspond.  \\
            \hline
             & PSGN 4D~\cite{fan2017point}  & - & 0.101 & 0.102 \\
             \multirow{2}*{S1} & ONet 4D~\cite{mescheder2019occupancy} & 77.9\% & 0.084 & - \\
             & OFlow~\cite{niemeyer2019occupancy}  & 81.5\% & 0.065 & 0.094 \\
             & Ours & \textbf{84.9\%} & \textbf{0.055} & \textbf{0.080} \\
             \hline
             \hline
             & PSGN 4D~\cite{fan2017point}  & - & 0.119 & 0.131 \\
             \multirow{2}*{S2} & ONet 4D~\cite{mescheder2019occupancy} & 66.6\% & 0.140 & - \\
             & OFlow~\cite{niemeyer2019occupancy}  & 72.3\% & 0.084 & 0.117 \\
              & Ours & \textbf{76.2\%} & \textbf{0.071} & \textbf{0.098} \\
             \bottomrule
            \end{tabular}
        \vspace{-5pt}
        \caption{Quantitative comparisons on the task of \textbf{4D Shape Reconstruction} from \textbf{time-evenly} sampled point cloud sequences. We evaluate the performance on the \emph{test} set of (S1) \textbf{unseen motions} (but seen individuals)  and (S2) \textbf{unseen individuals}. The metrics of Chamfer distance, correspondence, and IoU are reported.}
        \label{tab:AE:even}
        \vspace{-0pt}
    \end{table}
    
    
    \begin{table}[!t]
    \vspace{-5pt}
    \renewcommand\arraystretch{1.2}
    \setlength{\tabcolsep}{2mm}
        \centering
        \begin{tabular}{c | c | c c c}
        \toprule 
         & Method &  IoU & Chamfer & Correspond.  \\
        \hline
         & PSGN 4D~\cite{fan2017point}  & - & 0.148 & 0.121\\
         \multirow{2}*{S1} & ONet 4D~\cite{mescheder2019occupancy} & 71.9\% & 0.114 & - \\
         & OFlow~\cite{niemeyer2019occupancy}  & 76.9\% & 0.090  & 0.134 \\ 
         & Ours & \textbf{83.8\%} & \textbf{0.059} & \textbf{0.090} \\
        \hline
        \hline
         & PSGN 4D~\cite{fan2017point}  & - & 0.155 & 0.140 \\
         \multirow{2}*{S2} & ONet 4D~\cite{mescheder2019occupancy} & 62.8\% & 0.130 & - \\
         & OFlow~\cite{niemeyer2019occupancy} & 67.2\% & 0.112 & 0.178 \\
          & Ours & \textbf{74.8\%} & \textbf{0.076} & \textbf{0.118} \\
         \bottomrule
        \end{tabular}
        \vspace{-5pt}
        \caption{Quantitative comparisons on the task of \textbf{4D Shape Reconstruction} from \textbf{time-unevenly} sampled point cloud sequence with large variations between adjacent frames.} 
    \label{tab:AE:uneven}
    \vspace{-10pt}
    \end{table}
        \begin{figure}[t]
            \vspace{-5pt}
             \centering            
             \includegraphics[scale=0.5]{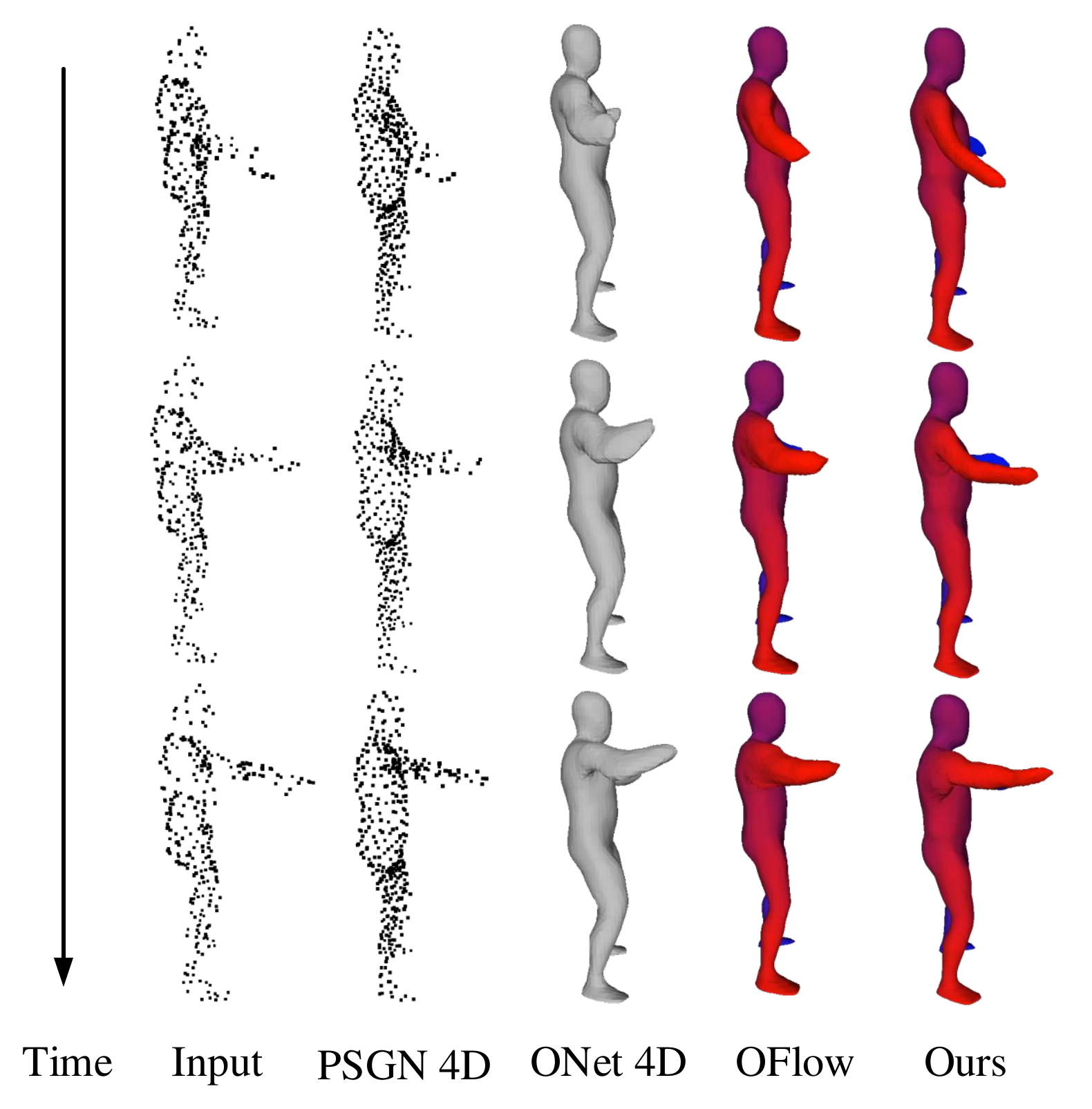}
             \vspace{-6pt}
             \caption{\textbf{Qualitative Results on 4D Shape Reconstruction.} We qualitatively show the input of three unequally space and time steps with large variations, and the output of PSGN 4D~\cite{fan2017point}, ONet 4D~\cite{mescheder2019occupancy}, and OFlow~\cite{niemeyer2019occupancy}. {Colors ranging from red to blue index mesh faces to better illustrate the surface correspondence across time.}}
            \label{fig:ae}
            \vspace{-16.8pt}
       \end{figure}
       We first compare the performance of our approach with
       previous methods for reconstructing time-varying surfaces from the sparse point cloud sequences in two kinds of input, including time-evenly and time-unevenly sampled sequences. The network input is 300 discrete point trajectories randomly sampled from dynamic groundtruth surfaces. In order to simulate the noises in the real world, we add gaussian noise with standard deviation 0.05 to perturb the point clouds.
       {For the former one, each trajectory consists of $K = 17$ time steps with uniform intervals. For the latter one, we randomly select 6 frames from a long segment of 50 time steps as input. Each trajectory experiences non-uniform intervals and large variations.}
       
       The quantitative and qualitative comparisons are respectively shown in Table~\ref{tab:AE:even} Table~\ref{tab:AE:uneven} and Fig.~\ref{fig:ae}.  As shown in the quantitative results, our approach achieves superior performance over all previous methods. 
       From the Fig.~\ref{fig:ae}, we can observe that our method can capture plausible motions and correspondences over time but ONet 4D can not. PSGN 4D predicts sparse and noisy point cloud sequences, causing the challenge to get clean dynamic surface meshes. {Besides, compared to OFlow~\cite{niemeyer2019occupancy}, our method can achieve more robust geometry recoveries benefiting from the proposed multi-frame shape information aggregation. Moreover, large performance improvements shown in Table~\ref{tab:AE:uneven} demonstrate that our multi-frame bundled correspondence modeling can achieve higher robustness on non-uniform sequences with large variations.}
    
    \subsection{4D Shape Completion}
    In addition to 4D shape reconstruction experiments, we also compare the performance on incomplete observations. Specifically, we create partial point clouds by randomly select 5 seeds on the surface and discard those regions within the radius of 0.1. The input is a sequence of $K = 6$ incomplete point clouds randomly sampled from a long segment, and each contains 300 points.  The quantitative and qualitative results are respectively shown in Table~\ref{tab:ablation_completion} and Fig.~\ref{fig:ablation}. The better performances verify the superiority of our approach in the dynamic surface recoveries with temporal coherence from incomplete observations.
    
    \begin{table}[t]
\vspace{-5pt}
\renewcommand\arraystretch{1.2}
\setlength{\tabcolsep}{2mm}
        \centering
        \begin{tabular}{c | c | c c c}
        \toprule
         & Method &  IoU & Chamfer & Correspond.  \\
        \hline
          & OFlow       &  75.9\%  & 0.094  & 0.142  \\
          & Ours (C1)   &  81.0\%  & 0.070  & 0.112\\
     S1   & Ours (C2)   &  58.6\%  & 0.124  & 0.254 \\
          & Ours (Full) & \textbf{82.4\%} & \textbf{0.064}  & \textbf{0.105} \\

         \hline\hline
          & OFlow      & 67.0\% & 0.113  & 0.183  \\
          & Ours (C1)  & 71.7\%  & 0.087  & 0.139  \\
     S2   & Ours (C2)  & 56.8\%   & 0.164  & 0.344  \\
          & Ours (Full) & \textbf{72.9\%} & \textbf{0.082} & \textbf{0.134} \\
         \bottomrule
        \end{tabular}
        \vspace{-5pt}
        \caption{\textbf{Ablation studies \& 4D Shape Completion}: Ours (C1) indicates our method without using the proposed spatio-temporal encoder, and Ours (C2) denotes our method without using the parallel correspondence modeling. }
    \label{tab:ablation_completion}
    \vspace{-10pt}
\end{table}
    
    \subsection{Ablation Studies}
\begin{figure}[!t]
        \vspace{-5pt}
            \centering
            \includegraphics[scale=0.62]{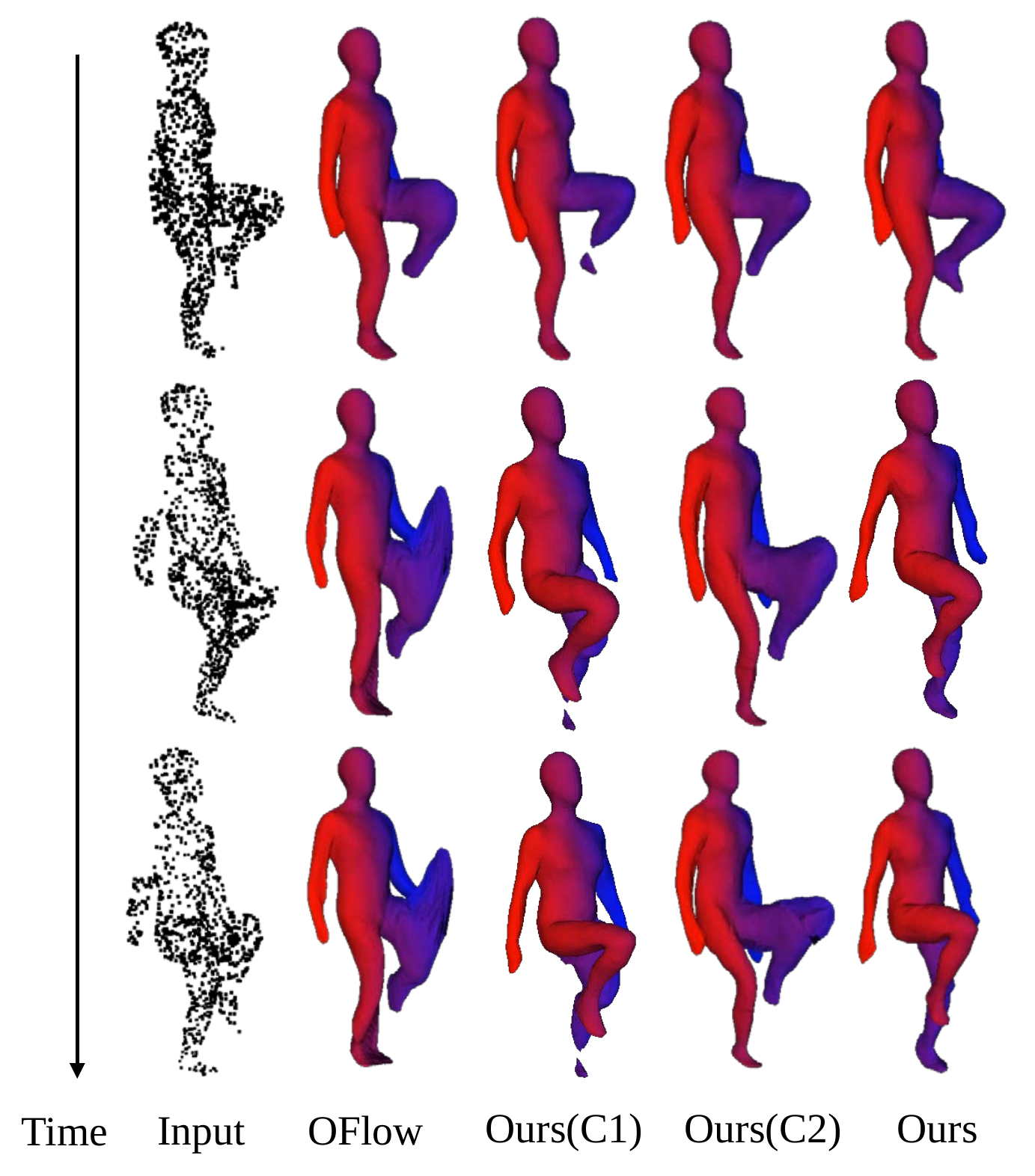}
           \vspace{-15pt} \caption{\textbf{4D Shape Completion \& Ablation Studies}: Ours (C1) indicates our method without using the proposed spatio-temporal encoder, and Ours (C2) denotes our method without using the parallel correspondence modeling.}
          \label{fig:ablation}
          \vspace{-5pt}
    \end{figure}

    {Our whole framework contains two key modules. In this
    section, we conduct the ablation studies by alternatively removing one of them to verify the necessity of each module. We perform experiments on the task of 4D shape completion from non-uniform sequences with large variations.
    
    \noindent \textbf{Without spatio-temporal encoder (C1)} Based on our pipeline, an alternative solution to learn the spatio-temporal representations is to individually utilize {Res-PointNet~\cite{charles2017pointnet} (details described in the supplementary material)} to process the 4D point cloud with time information at each frame.   The comparison results are shown in Fig.~\ref{fig:ablation} and Table~\ref{tab:ablation_completion}. As can be seen, our method achieves more complete geometry as the designed spatio-temporal encoder can efficiently aggregate the shape properties of all frames.
     
    \noindent \textbf{Without parallel correspondence modeling (C2)}
     Instead of parallelly modeling the dense correspondences, an alternative solution is to predict the occupancy field transformations between adjacent frames. As shown in Fig.~\ref{fig:ablation}, the sequential correspondence modeling accumulates the prediction errors in the legs in the second frame, resulting in wrong shape predictions in subsequent frames and non-continuous human dynamics. We also verify it by the increased correspondence distance (see Table~\ref{tab:ablation_completion}). Moreover, the sequential manner adopted by OFlow and Ours (C2) easily produces distorted results such as stretched surfaces around the left knee when capturing large human variations.}
    
    \subsection{Space and Time Complexity}
    We compare our method to OFlow~\cite{niemeyer2019occupancy} in terms of memory footprint and computational efficiency.
    We train both models using a batch size of 16 for 4D shape reconstruction from a sequence of 17 time steps with uniform intervals and report the training memory footprint, total training time.  And we calculate the average of batch training time, batch forward time, and batch backward time in the initial 10k iterations of training. We also report the average network inference time using a batch size of 1 for 1k test sequences.
    Both models were run on a single GTX 1080 Ti. We observed the slow training procedure of OFlow~\cite{niemeyer2019occupancy} as ODE-solver requires demanding computations and gradually increases the number of iterations to meet the error tolerance. 
    From the results shown in Table~\ref{tab:complexity}, we can see that although our model has a higher training memory footprint, it is about 4 times faster in training and 8 times in inference.
    
    \subsection{Shape Matching}
    {In this section, we investigate our pipeline for the task of shape matching. The inputs are the underlying surfaces of two randomly sampled point clouds, and the outputs are the point displacements of source surface to the target surface. Since this task does not need to recover 3D surface meshes, the model consists of only a spatio-temporal encoder and a dense correspondence decoder, and the training is only supervised by the correspondence loss in Eq.~\ref{EqnCorrLoss}. From the quantitative comparisons shown in Table~\ref{tab:shapematching}, we can conclude that although our method is primarily designed for 4D reconstruction, it can also predict accurate correspondences. Moreover, we remark that our inference speed is remarkably faster than that of CPD~\cite{myronenko2010point} and OFlow~\cite{niemeyer2019occupancy}, with approximately 31,000 and 30 times faster, respectively.
    }
    
        \begin{table}[!t]
        \vspace{-5pt}
        \renewcommand\arraystretch{0.99}
        \setlength{\tabcolsep}{1.2mm}
            \centering
            \begin{tabular}{c c c c c}
            \toprule
             Method & Mem. (GB) & Train (day) & Inference (s)  \\
            \hline             OFlow~\cite{niemeyer2019occupancy}  & 3.53  & 42 & 1.84 \\
             Ours     & 10.8  & 10  &  0.23 \\ 
             \hline
             \hline
              & Forward (s). & Backward (s). & Train (s) \\
            \hline
             OFlow~\cite{niemeyer2019occupancy} & 0.33 & 4.02  & 4.35 \\
             Ours & 0.45 & 0.49 & 0.94 \\
             \bottomrule
            \end{tabular}
            \vspace{-5pt}
            \caption{\textbf{Space and time complexity comparison} between OFlow~\cite{niemeyer2019occupancy} and our method.}
        \label{tab:complexity}
    \end{table}

    \begin{table}[!t]
    	\renewcommand\arraystretch{0.99}
        \setlength{\tabcolsep}{1.2mm}
        	\centering
            \begin{tabular}{c|*{2}{c}}
        		\toprule
        		Method  &  Correspond & Time(s) \\ 
        		\hline
        		Nearest Neighbor & 0.374  &\textbf{0.004} \\
        		Coherent Point Drift~\cite{myronenko2010point}   & 0.189  & 343.6   \\
        		OFlow~\cite{niemeyer2019occupancy} &0.167 & 0.309 \\
        		Ours   & \textbf{0.102} & \textbf{0.011}  \\
        		\bottomrule
        	\end{tabular}
        	\vspace{-5pt}
        	\caption{\textbf{Shape Matching Experiments.}~We report the correspondence distance (correspond) of two randomly sampled point clouds with the size 10k.}
        \label{tab:shapematching}
        \vspace{-12pt}
    \end{table}
\section{Conclusion}
\label{SecConclu}

We have proposed a novel learning framework to reconstruct time-varying surfaces from point cloud sequences. The overall framework includes a flexible framework for learning robust spatio-temporal shape representations from 4D point clouds and an efficient cross-frame correspondence decoder that simultaneously models the occupancy field transformations from the first frame to others. Comparisons with previous works demonstrate that our approach can achieve more accurate geometries, better temporal continuity while significantly improves the computation efficiency. One limitation of our method is that to achieve superior practical efficacy and efficiency, we sacrifice theoretically temporal continuity due to the discrete field transformation, which will be further explored in the future works.
\vspace{-2pt}
\par\noindent\textbf{Acknowledgement.} 
This work was partially supported by the National Natural Science Foundation of China (No.:~61771201), the Program for Guangdong Introducing Innovative and Entrepreneurial Teams (No.: 2017ZT07X183), the Guangdong R$\&$D key project of China (No.: 2019B010155001), Alibaba DAMO Academy, and Hong Kong RGC GRF (No.: 15221618).

\newpage{
\section*{Appendix}
    \begin{figure}[!t]
        \centering
        \includegraphics[width=0.85\linewidth]{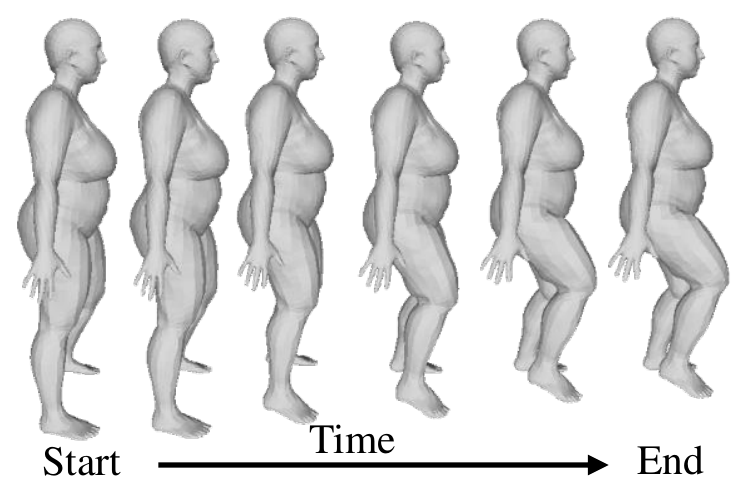}
        \vspace{10pt}
    	\includegraphics[width=0.90\linewidth]{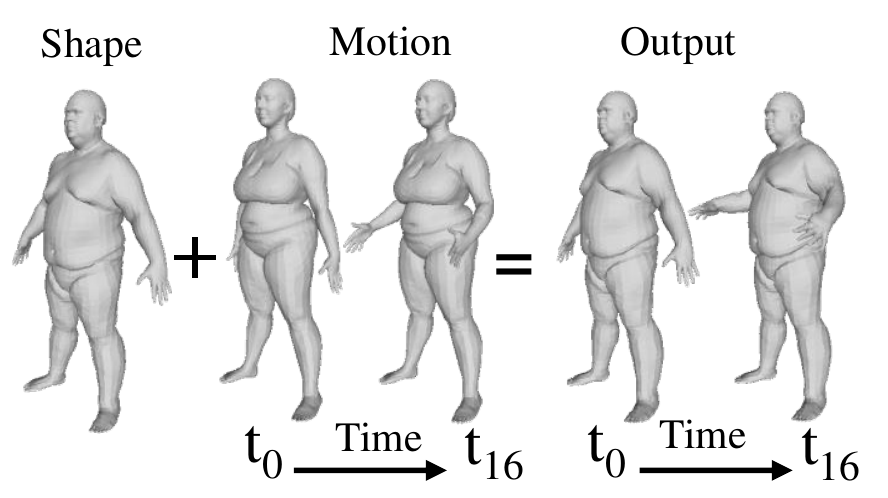}
        \vspace{-5pt}
        \caption{Qualitative results on Shape Interpolation (top) and Motion Transfer (bottom).}
        \label{fig:interpolation}
        \vspace{-10pt}
    \end{figure}
  
In this {supplementary material}, we first elaborate the details of network architectures in Sec.~\ref{SuppleNet}, and then provide more detailed qualitative and quantitative comparisons against previous works in Sec.~\ref{SuppleAddRes}, including PSGN 4D~\cite{fan2017point}, ONet 4D~\cite{mescheder2019occupancy}, and OFlow~\cite{niemeyer2019occupancy}. Finally, we investigate our pipeline for other applications such as shape interpolation and motion transfer in Sec.~\ref{SuppleOtherApp}.

\section{Network Architectures}
\label{SuppleNet}
    \subsection{Spatio-temporal Encoder}
    The input of a 4D spatio-temporal point cloud is donated by $\mathcal{S} = \{ {\mathcal{S}_k} \}_{k=0}^{K-1}$, where ${ \mathcal{S}_k } = \{ \mathbf{s}_i \in \mathbb{R}^4 = \{ x_i, y_i, z_i, t_k \}_{i=0}^{{N_k}-1} \}$. ${N_k}$ is the number of points at frame $k \in [0, K-1]$ and at time $t_k \in [{t_0}, {t_{K-1}}] \subset \mathbb{R}$ with ${N} = \sum_{k=1}^K N_k$. To aggregate shape properties and explore dynamics variations from $\mathcal{S}$, we design a novel spatio-temporal encoder, whose network architecture is composed of S-PointNet, T-PointNet, and FusionNet. Both S-PointNet and T-PointNet are implemented by ResNet~\cite{he2016deep} variants of PointNet~\cite{charles2017pointnet}, which is shown in Fig.~\ref{fig:poinet}. The only difference between them is that the input dimension $d=3$ and the number of input points $M = N_k$ in S-PointNet, whereas $d=4$ and $M = N$ in T-PointNet due to the use of time information. Specifically, the T-PointNet accepts and transforms the entire 4D point cloud $\mathcal{S} = \{ {\mathcal{S}_k} \}_{k=0}^{K-1}$ to extract a temporal representation $\mathbf{z}_t$. The S-PointNet produces a sequence of geometric representations $\mathbf{z}_{s,0}, \mathbf{z}_{s,1}, ..., \mathbf{z}_{s,K-1}$ by individually processing each frame $\mathcal{S}_k$ without considering timestamps. Finally, the temporal representation $\mathbf{z}_t$ is replicated and concatenated with geometric representations $\mathbf{z}_{s,0}, \mathbf{z}_{s,1}, ..., \mathbf{z}_{s,K-1}$, and then the new formed features are passed through the FusionNet to obtain a sequence of 128-dimensional descriptors $\{ \mathbf{z}_0, \mathbf{z}_1, ... \mathbf{z}_{K-1} \} $. The FusionNet is implemented as FC-256 $\rightarrow$ FC-128. 
    
    \begin{figure*}[h]
        \centering
        \includegraphics[scale=0.6]{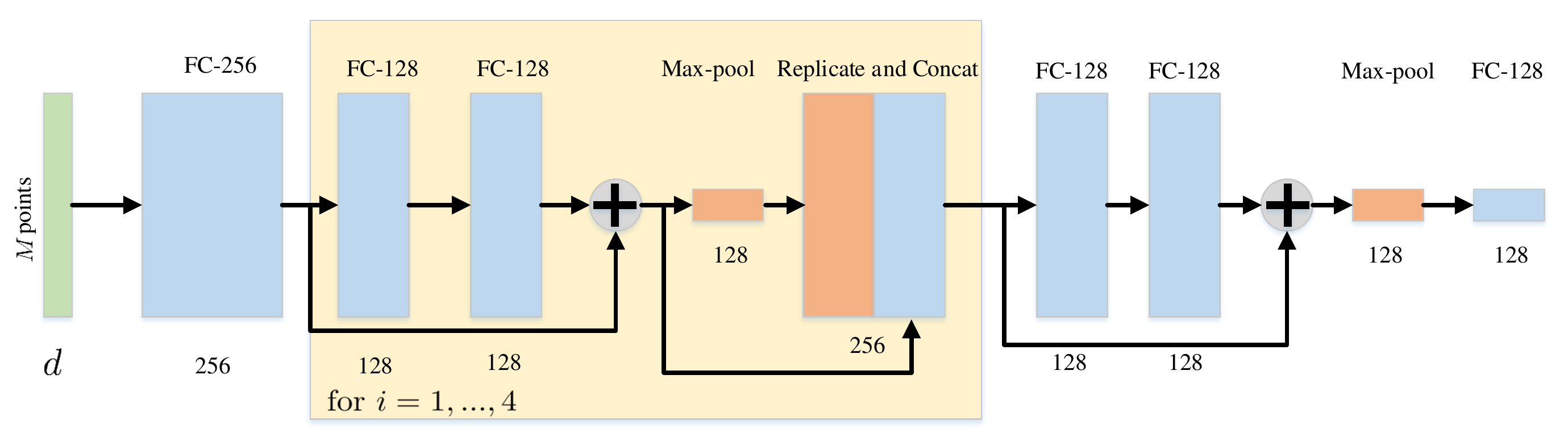}
        \caption{\textbf{ResNet~\cite{he2016deep} variants of PointNet}~\cite{charles2017pointnet}. The input $d-$dimensional point cloud ($d=$ 3 or 4 depending on whether incorporating time information) goes through four ResNet-FC blocks with skip connections and max-pooling operations and feature replications. Then we pass the feature produced by the last ResNet-FC block through a max-pooling and a fully connected layer to obtain a 128-dimensional feature vector.}
         \label{fig:poinet}
    \end{figure*}
    
    \subsection{Occupancy and Correspondence Decoders}
    \begin{figure*}[h]
        \centering
        \includegraphics[scale=0.55]{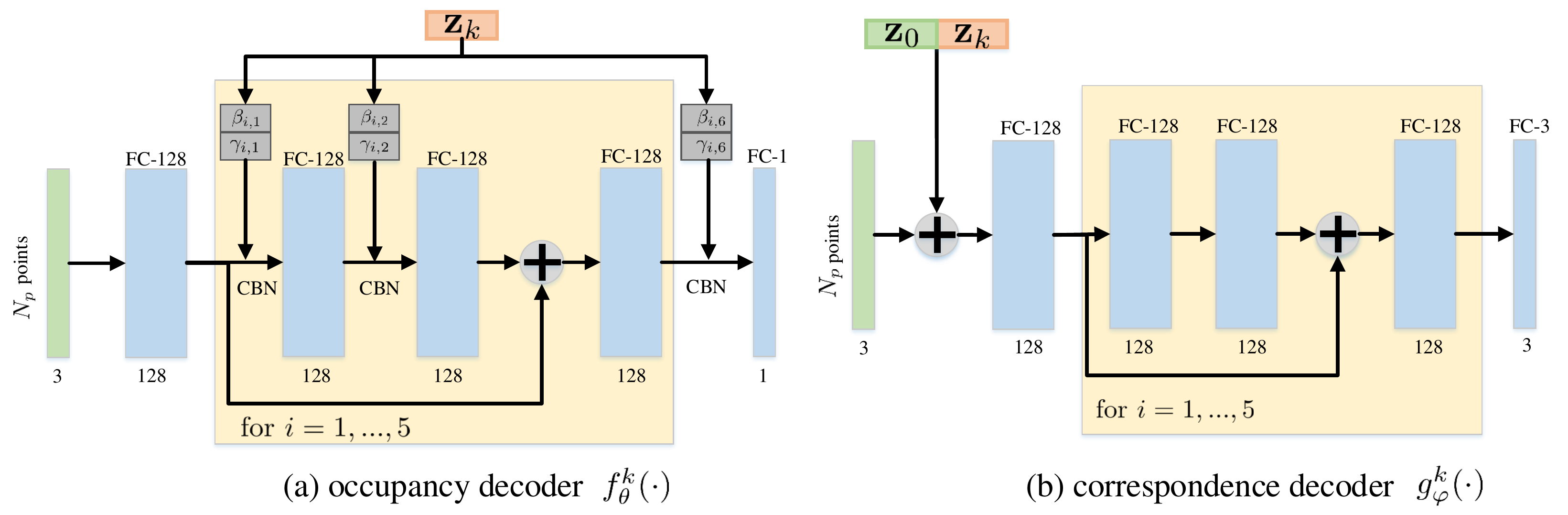}
        \caption{\textbf{Network architectures for occpancy and correspondence decoders}. The occupancy field $\mathbf{O}_k$ at time $t_k$ is predicted by $f_\theta^{k}(\cdot)$ conditioned on the spatio-temporal representation $\mathbf{z}_k$. The dense correspondences between $\mathbf{O}_0$ and $\mathbf{O}_k$ are predicted by $g_\varphi^{k}(\cdot)$ conditioned on the concatenation of associate representations $\mathbf{z}_0$ and $\mathbf{z}_k$.}
        \label{fig:decoder_net}
    \end{figure*}
    
    In this section, we explain the network architectures for the occupancy decoder and dense correspondence decoder. The occupancy decoder predicts the occupancy probabilities for the sampled $N_p$ points at time $t_k$. The dense correspondence decoder
    outputs the displacements from $\mathbf{p}_0$ at time $t_0$ to $\mathbf{p}_k$ located in the coordinate system at time $t_k$.
    As shown in Fig.~\ref{fig:decoder_net}, both two networks cascade five ResNet-FC blocks, and each block is composed of two fully connected layers with skip connections. Similar to~\cite{mescheder2019occupancy}, we use the conditional batch normalization (CBN)~\cite{de2017modulating, dumoulin2016adversarially} for the occupancy network.
    
    \setlength{\tabcolsep}{3pt}
\begin{table*}[t]
    \vspace{5pt}
	\renewcommand\arraystretch{1.2}
	\begin{center}
		\begin{tabular}{*{14}{c}}
			\toprule
			\multirow{2}*{Time step} & \multicolumn{4}{c}{IoU}   & \multicolumn{4}{c}{Chamfer} & \multicolumn{4}{c}{Correspond.}\\
			\cmidrule(lr){2-5} \cmidrule(lr){6-9} \cmidrule(lr){10-13}
			& PSGN 4D &  ONet 4D & OFlow  & Ours  & PSGN 4D &  ONet 4D & OFlow  & Ours  & PSGN 4D &  ONet 4D & OFlow  & Ours \\
			\midrule
			\midrule
           0     & -  & 76.3\% & 83.1\%  & \textbf{85.6\%}
                 & 0.104 & 0.090 & 0.059  & \textbf{0.052}
                 & 0.075  & -  & 0.057  & \textbf{0.047}\\
           
           1     & -  & 77.1\% & 83.1\%   & \textbf{85.5\%}
                 & 0.103 & 0.087 & 0.059 & \textbf{0.053}
                 & 0.073 & -    & 0.062  & \textbf{0.053}\\
                   
           2     & -  & 77.5\% & 82.8\%  & \textbf{85.4\%}
                 & 0.102 & 0.084 & 0.061 & \textbf{0.053}
                 & 0.081 & - &0.069  & \textbf{0.059}\\
                   
           3     & -  & 77.8\% & 82.5\%   & \textbf{85.3\%}
                 & 0.102  & 0.084 & 0.061  & \textbf{0.053}
                 & 0.090  & -  & 0.077  & \textbf{0.064}\\
                
           4     & -  & 78.0\% & 82.2\%   & \textbf{85.0\%}
                 & 0.106  & 0.083  & 0.062  & \textbf{0.054}
                 & 0.097  & -  & 0.083  & \textbf{0.070}\\
                 
           5     & -  & 78.2\% & 82.0\% & \textbf{85.1\%}
                 & 0.106  & 0.082  & 0.063  & \textbf{0.054}
                 & 0.102  & - & 0.088  & \textbf{0.074}\\
                   
           6     & -  & 78.3\% & 81.8\% & \textbf{85.1\%}
                 & 0.106  & 0.082  & 0.064  & \textbf{0.054}
                 & 0.106  & - & 0.092  & \textbf{0.078}\\
                   
           7     & -  & 78.4\% & 81.6\%  & \textbf{85.1\%}
                 & 0.106  & 0.081  & 0.064  & \textbf{0.054}
                 & 0.109  & - & 0.095  & \textbf{0.082}\\
                   
           8     & -  & 78.5\% & 81.4\%  & \textbf{85.0\%}
                 & 0.106  & 0.081  & 0.065  & \textbf{0.055}
                 & 0.111  & - & 0.098  & \textbf{0.085}\\
                   
           9     & -  & 78.5\% & 81.3\% & \textbf{85.0\%}
                 & 0.107  & 0.081  & 0.066  & \textbf{0.055}
                 & 0.111  & - & 0.101 & \textbf{0.0888}\\
        
           10    & -  & 78.4\% & 81.1\% & \textbf{84.6\%}
                 & 0.107  & 0.082  & 0.066  & \textbf{0.055}
                 & 0.112  & - & 0.103 & \textbf{0.090}\\
           
           11    & -  & 78.4\% & 81.0\% & \textbf{84.6\%}
                 & 0.107  & 0.082  & 0.067  & \textbf{0.055}
                 & 0.112  & - & 0.106 & \textbf{0.092}\\
                   
           12    & -  & 78.2\% & 80.8\% & \textbf{84.6\%}
                 & 0.108  & 0.083  & 0.068  & \textbf{0.056}
                 & 0.112  & - & 0.108 & \textbf{0.094}\\
                   
           13    & -  & 78.1\% & 80.6\%  & \textbf{84.4\%}
                 & 0.108  & 0.083  & 0.069   & \textbf{0.056}
                 & 0.112  & - & 0.111 & \textbf{0.095}\\
            
           14    & -  & 77.8\% & 80.3\%  & \textbf{84.3\%}
                 & 0.109  & 0.085  & 0.070  & \textbf{0.056}
                 & 0.113  & - & 0.114 & \textbf{0.096}\\
                 
           15    & -  & 77.4\% & 80.0\% & \textbf{84.3\%}
                 & 0.110  & 0.086  & 0.071  & \textbf{0.056}
                 & 0.115  & - & 0.119 &\textbf{0.097}\\
           
           16   & -  & 76.9\%  & 79.5\% & \textbf{84.3\%}
                 & 0.113  & 0.089  & 0.073  & \textbf{0.056}
                 & 0.120  & - & 0.125 & \textbf{0.098}\\
           
        \midrule
            mean & -  & 77.9\% & 81.5\% & \textbf{84.9\%}
                 & 0.108  & 0.084  & 0.065  & \textbf{0.055}
                 & 0.102  & - & 0.094 & \textbf{0.080}\\
        \bottomrule
        \end{tabular}
        \caption{ Quantitative comparisons on the task of \textbf{4D shape reconstruction} from \textbf{time-evenly} sampled point cloud sequences. The metrics of IoU, Chamfer distance, and Correspondence distance for all 17 time steps for the \emph{test} set (S1) \textbf{unseen motions (but seen individuals)} are reported. }
        \label{Tab:ae:motion}
        \end{center}
        \vspace{5pt}
\end{table*}
    \setlength{\tabcolsep}{3pt}
\begin{table*}[t]
    \vspace{5pt}
	\renewcommand\arraystretch{1.2}
	\begin{center}
		\begin{tabular}{*{14}{c}}
			\toprule
			\multirow{2}*{Time step} & \multicolumn{4}{c}{IoU}   & \multicolumn{4}{c}{Chamfer} & \multicolumn{4}{c}{Correspond.}\\
			\cmidrule(lr){2-5} \cmidrule(lr){6-9} \cmidrule(lr){10-13}
			& PSGN 4D &  ONet 4D & OFlow  & Ours  & PSGN 4D &  ONet 4D & OFlow  & Ours  & PSGN 4D &  ONet 4D & OFlow  & Ours \\
			\midrule
			\midrule
           0   & -     & 72.2\% & 79.6\%  & \textbf{84.9\%}
               & 0.150 & 0.113  & 0.070  & \textbf{0.056}
               & 0.079 & -      & 0.069  & \textbf{0.051}\\
           
           1  & -      & 72.8\%  & 78.3\%  & \textbf{84.0\%}
              & 0.148  & 0.110   & 0.088  & \textbf{0.058}
              & 0.107  & -       & 0.130  & \textbf{0.094}\\
                   
           2  & -      & 71.8\%  & 77.6\%  & \textbf{83.5\%}
              & 0.147  & 0.113   & 0.090  & \textbf{0.059}
              & 0.122  & -       & 0.137  & \textbf{0.097}\\
                   
           3  & -      & 71.6\% & 75.9\%  & \textbf{83.6\%}
              & 0.147  & 0.114  & 0.093  & \textbf{0.060}
              & 0.137  & -      &  0.145 & \textbf{0.093}\\
                
           4  & -      & 71.6\%  & 75.3\%  & \textbf{83.4\%}
              & 0.148  & 0.115  & 0.096  & \textbf{0.060}
              & 0.140  & -      & 0.160  & \textbf{0.100}\\
                 
           5  & -      & 71.6\% &  74.7\% & \textbf{83.8\%}
              & 0.149  & 0.116  & 0.098  & \textbf{0.060}
              & 0.144  & -      & 0.166 & \textbf{0.102}\\

        \midrule
            mean & -      & 71.9\% & 76.9\% & \textbf{83.8\%}
                 & 0.148  & 0.114  & 0.090  & \textbf{0.059}
                 & 0.121  & -      & 0.134  & \textbf{0.090}\\
        \bottomrule
        \end{tabular}
        \caption{ Quantitative comparisons on the task of \textbf{4D Shape Reconstruction} from \textbf{time-unevenly} sampled point cloud sequences with large variations between adjacent frames. The metrics of IoU, Chamfer distance, and Correspondence distance for all 6 time steps for the \emph{test} set (S1) \textbf{unseen motions (but seen individuals)} are reported.
        }
       \label{Tab:ae_uneven:motion}
        \end{center}
        \vspace{5pt}
\end{table*}

    \setlength{\tabcolsep}{3pt}
\begin{table*}[t]
    \vspace{5pt}
	\renewcommand\arraystretch{1.2}
	\begin{center}
		\begin{tabular}{*{14}{c}}
			\toprule
			\multirow{2}*{Time step} & \multicolumn{4}{c}{IoU}   & \multicolumn{4}{c}{Chamfer} & \multicolumn{4}{c}{Correspond.}\\
			\cmidrule(lr){2-5} \cmidrule(lr){6-9} \cmidrule(lr){10-13}
			& PSGN 4D &  ONet 4D & OFlow  & Ours  & PSGN 4D &  ONet 4D & OFlow  & Ours  & PSGN 4D &  ONet 4D & OFlow  & Ours \\
			\midrule
			\midrule
            0    & -      & 64.8\%  & 74.2\%  & \textbf{76.8\%}
                 & 0.121  & 0.148  & 0.077  & \textbf{0.068}
                 & 0.093  & -  & 0.077  & \textbf{0.065}\\
           
            1    & -    & 65.7\% & 74.1\%   & \textbf{76.8\%}
                 & 0.120  & 0.143 & 0.077  & \textbf{0.069}
                 & 0.098  & -  & 0.082  & \textbf{0.071}\\
                   
            2    & -     & 66.2\%  & 73.8\%   & \textbf{76.7\%}
                 & 0.119  & 0.141 & 0.078  & \textbf{0.069}
                 & 0.108  & -  & 0.089  & \textbf{0.075}\\
                   
            3    & -      & 66.5\%  & 73.4\%  & \textbf{76.5\%}
                 & 0.118  & 0.139   & 0.079 & \textbf{0.069}
                 & 0.117  & -  & 0.096  & \textbf{0.080}\\
                
            4    & -      & 66.7\%  & 73.0\%    & \textbf{76.5\%}
                 & 0.118  & 0.138 &  0.081  & \textbf{0.070}
                 & 0.127  & -  & 0.102  & \textbf{0.085}\\
                 
            5    & -  & 66.9\% & 72.7\%  & \textbf{76.6\%}
                 & 0.118  & 0.137 & 0.082  & \textbf{0.070}
                 & 0.135  & -  & 0.108  & \textbf{0.089}\\
                   
            6    & -  & 67.0\%  & 72.4\%  & \textbf{76.3\%}
                 & 0.119  & 0.137 & 0.083   & \textbf{0.070}
                 & 0.143  & -  & 0.113  & \textbf{0.094}\\
                   
            7    & -  & 67.1\% & 72.2\%  & \textbf{76.2\%}
                 & 0.119  &  0.136 & 0.084  & \textbf{0.071}
                 & 0.149  & -  & 0.117  & \textbf{0.098}\\
                   
            8    & -  & 67.2\% & 72.0\% & \textbf{76.0\%}
                 & 0.119  & 0.136 & 0.085  & \textbf{0.071}
                 & 0.155  & -  & 0.121  & \textbf{0.101}\\
                   
            9    & -  & 67.3\% & 71.9\%  & \textbf{76.1\%}
                 & 0.119  & 0.136 & 0.085  & \textbf{0.071}
                 & 0.160  & -  & 0.124  & \textbf{0.104}\\
        
            10   & -  & 67.3\% & 71.8\%  & \textbf{76.1\%}
                 & 0.119  & 0.136 & 0.086  & \textbf{0.072}
                 & 0.164  & -  & 0.127  & \textbf{0.107}\\
           
            11   & -  & 67.2\% & 71.7\%  & \textbf{75.9\%}
                 & 0.119  & 0.137 & 0.086  & \textbf{0.072}
                 & 0.168  & -  & 0.130  & \textbf{0.109}\\
                   
            12   & -  & 67.1\% & 71.5\%  & \textbf{75.9\%}
                 & 0.119  & 0.138 & 0.087  & \textbf{0.072}
                 & 0.172  & -  & 0.133  & \textbf{0.112}\\
                   
            13   & -  & 66.9\% & 71.4\%  & \textbf{75.8\%}
                 & 0.119  & 0.139 & 0.087  & \textbf{0.072}
                 & 0.176  & -  &  0.136 & \textbf{0.114}\\
            
            14   & -  & 66.6\% & 71.3\%  & \textbf{75.7\%}
                 & 0.119  & 0.141 & 0.088  & \textbf{0.073}
                 & 0.181  & -  & 0.140  & \textbf{0.116}\\
           
            15   & -  & 66.3\% & 71.0\%  & \textbf{75.6\%}
                 & 0.120  & 0.143 & 0.089  & \textbf{0.073}
                 & 0.187  & -  &  0.145 & \textbf{0.119}\\
           
            16   & -  & 65.8\% &  70.7\% & \textbf{75.7\%}
                 & 0.121  & 0.146 & 0.090  & \textbf{0.074}
                 & 0.195  & -  & 0.150  & \textbf{0.121}\\
           
        \midrule
            mean & -  & 66.6\% & 72.3\%  & \textbf{76.2\%}
                 & 0.119  & 0.140 & 0.084   & \textbf{0.071}
                 & 0.131  & -  & 0.117  & \textbf{0.098}\\
        \bottomrule
        \end{tabular}
        \caption{Quantitative comparisons on the task of \textbf{4D Shape Reconstruction} from \textbf{time-evenly} sampled point cloud sequences. The metrics of IoU, Chamfer distance, and Correspondence distance for all 17 time steps for the \emph{test} set (S2) \textbf{unseen individuals} are reported.}
         \label{Tab:ae:individual}
        \end{center}
        \vspace{5pt}
\end{table*}

    \setlength{\tabcolsep}{3pt}
\begin{table*}[t]
    \vspace{5pt}
	\renewcommand\arraystretch{1.2}
	\begin{center}
		\begin{tabular}{*{14}{c}}
			\toprule
			\multirow{2}*{Time step} & \multicolumn{4}{c}{IoU}   & \multicolumn{4}{c}{Chamfer} & \multicolumn{4}{c}{Correspond.}\\
			\cmidrule(lr){2-5} \cmidrule(lr){6-9} \cmidrule(lr){10-13}
			& PSGN 4D &  ONet 4D & OFlow  & Ours  & PSGN 4D &  ONet 4D & OFlow  & Ours  & PSGN 4D &  ONet 4D & OFlow  & Ours \\
			\midrule
			\midrule
           0   & -      & 62.3\%  & 71.2\%  & \textbf{75.5\%}
               & 0.156  & 0.132  & 0.091  & \textbf{0.073}
               & 0.088  & -  & 0.086  & \textbf{0.072}\\
                 
           1  & -      & 63.6\% & 68.3\%  & \textbf{75.2\%}
              & 0.155  & 0.126 & 0.104  & \textbf{0.075}
              & 0.119  & -  & 0.150  & \textbf{0.108}\\
                   
           2  & -      & 63.4\% & 67.6\%  & \textbf{75.0\%}
              & 0.154  & 0.126  & 0.110  & \textbf{0.076}
              & 0.138  & -  & 0.173  & \textbf{0.120}\\
                   
           3 & -      & 62.8\% & 66.3\%  & \textbf{74.6\%}
             & 0.154  & 0.129  & 0.116 & \textbf{0.076}
             & 0.159  & -  &  0.195 & \textbf{0.122}\\
                   
           4  & -      & 62.4\%  & 64.9\%  & \textbf{74.2\%}
              & 0.154  & 0.131  & 0.125  & \textbf{0.078}
              & 0.165  & -  &  0.229 & \textbf{0.141}\\
           
           5   & -      & 62.4\%  & 64.6\% & \textbf{74.3\%}
               & 0.154  & 0.133 & 0.128  & \textbf{0.079}
               & 0.169  & -  &  0.238 & \textbf{0.144}\\
           
        \midrule
            mean & -      & 62.8\% &  67.2\% & \textbf{74.8\%}
                 & 0.155  & 0.130  &  0.112  & \textbf{0.076}
                 & 0.140  & -  & 0.178  & \textbf{0.118}\\
        \bottomrule
        \end{tabular}
        \caption{
        Quantitative comparisons on the task of \textbf{4D Shape Reconstruction} from \textbf{time-unevenly} sampled point cloud sequences with large variations between adjacent frames. The metrics of IoU, Chamfer distance, and Correspondence distance for all 6 time steps for the \emph{test} set (S2) \textbf{unseen individuals} are reported.
        }
         \label{Tab:ae_uneven:individual}
        \end{center}
        \vspace{5pt}
\end{table*}
    \begin{figure*}[!t]
            \centering
            \includegraphics[scale=0.65]{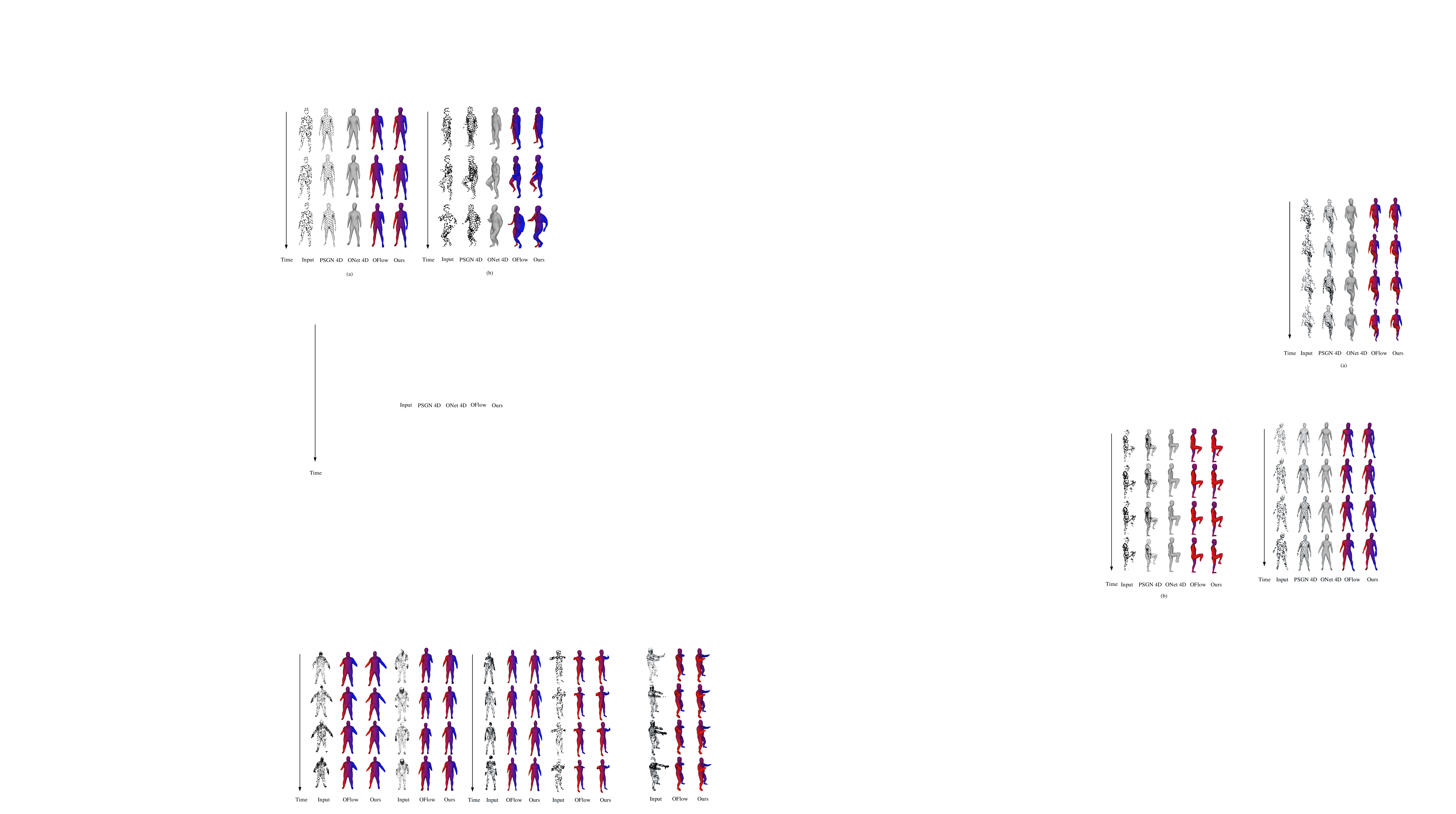}
            \vspace{-15pt}
            \caption{Qualitative comparisons on the \textbf{4D Shape Reconstruction} from \textbf{time-evenly} sampled point cloud (a) and \textbf{time-unevenly} sampled point cloud sequences (b).}
            \vspace{-10pt} \label{fig:supple_recon}
    \end{figure*}

    \begin{figure*}[!t]
            \centering
            \includegraphics[scale=0.53]{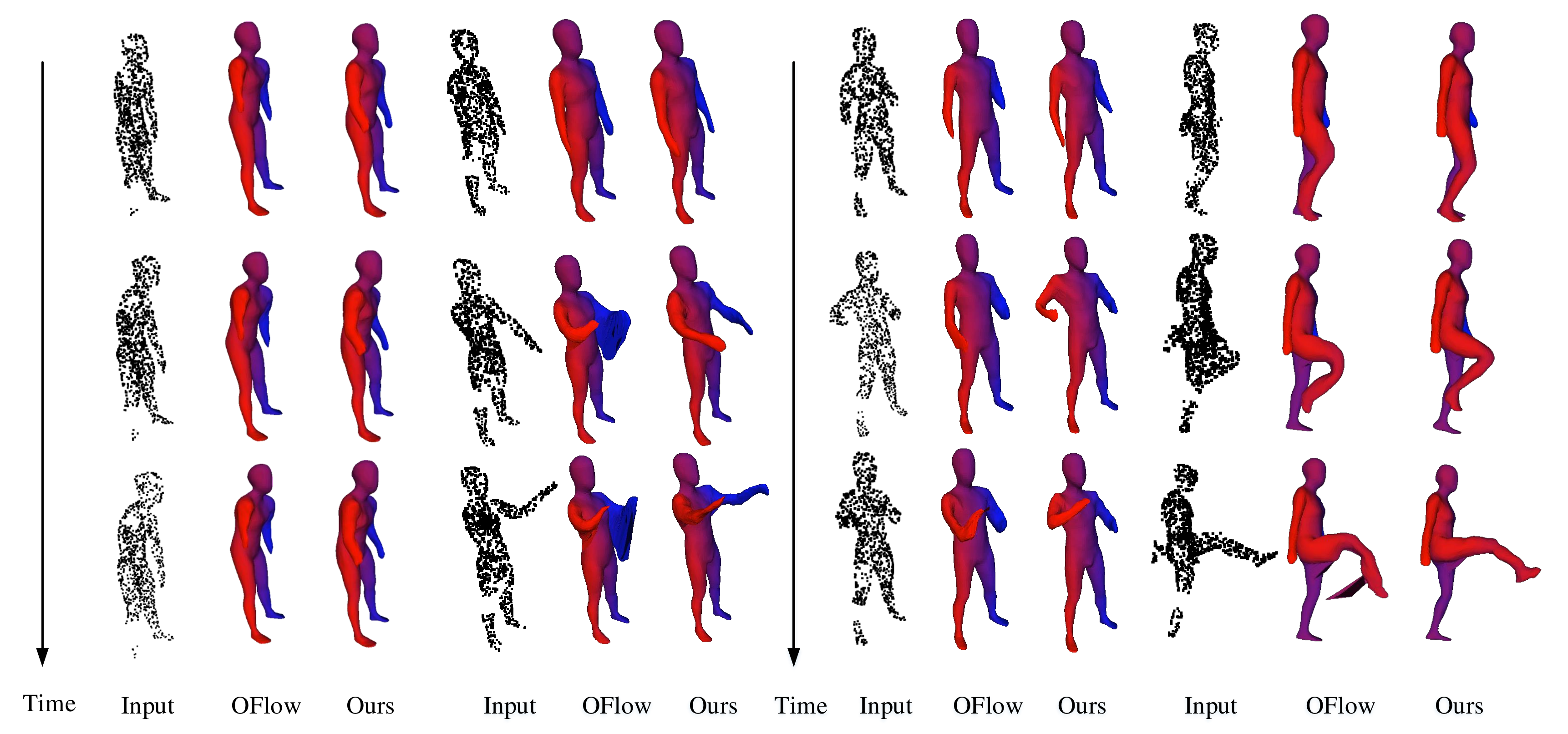}
            \vspace{-3pt}
            \caption{Qualitative comparisons on the task of \textbf{4D Shape Completion}.}
           \vspace{-10pt} \label{fig:supple_completion}
     \end{figure*}

\section{Additional Experiments} 
\label{SuppleExp}
    \subsection{Quantitative results}
    \label{SuppleAddRes}
    For the task of 4D shape reconstruction from time-evenly sampled point cloud sequences, quantitative comparisons for all 17 time steps  for all 6 time steps on the \emph{test} sets of (S1) \textbf{unseen motions (but seen individuals)} and (S2) \textbf{unseen individuals} are respectively shown in Table~\ref{Tab:ae:motion} and Table~\ref{Tab:ae:individual}.
    
    For the task of 4D shape reconstruction from time-unevenly sampled point cloud sequences with large variations between adjacent frames, quantitative comparisons for all 6 time steps on the \emph{test} sets of (S1) \textbf{unseen motions (but seen individuals)} and (S2) \textbf{unseen individuals} are respectively shown in Table~\ref{Tab:ae_uneven:motion} and Table~\ref{Tab:ae_uneven:individual}. And more qualitative comparisons are depicted in Fig.~\ref{fig:supple_recon}.
    
    For the task of 4D shape completion, quantitative and qualitative comparisons are respectively shown in Table~\ref{Tab:completion:motion}, Table~\ref{Tab:completion:individual}, and Fig.~\ref{fig:supple_completion}.

    \setlength{\tabcolsep}{3pt}
\begin{table*}[t]
    \vspace{-5pt}
	\renewcommand\arraystretch{1.2}
	\begin{center}
		\begin{tabular}{*{14}{c}}
			\toprule
			\multirow{2}*{Time step} & \multicolumn{4}{c}{IoU}   & \multicolumn{4}{c}{Chamfer} & \multicolumn{4}{c}{Correspond.}\\
			\cmidrule(lr){2-5} \cmidrule(lr){6-9} \cmidrule(lr){10-13}
			& OFlow &  C1 &  C2  & Ours  & OFlow &  C1 &  C2  & Ours  & OFlow &  C1 &  C2  & Ours \\
			\midrule
			\midrule
           0   & 79.1\%  & 81.8\%  & 81.5\%  & \textbf{83.6\%}
               & 0.080   & 0.068   & 0.066   & \textbf{0.060} 
               & 0.070   & 0.059  & 0.062   & \textbf{0.054}\\
           
           1   & 76.8\%  & 81.1\%  & 73.2\%  & \textbf{82.5\%}
               & 0.090   & 0.070   & 0.099   & \textbf{0.064} 
               & 0.134   & 0.120  & 0.218   & \textbf{0.112}\\
                   
           2   & 75.7\%  & 80.9\%  & 69.1\%   & \textbf{82.1\%}
               & 0.093   & 0.070   & 0.117   & \textbf{0.065} 
               & 0.148   & 0.120   & 0.258   & \textbf{0.115}\\
                   
           3   & 75.6\%  & 80.8\%  & 65.5\%   & \textbf{82.3\%}
               & 0.094   & 0.071   & 0.134   & \textbf{0.065} 
               & 0.151   & 0.118   & 0.272   & \textbf{0.111}\\
                
           4   & 74.3\%  & 80.9\%  & 60.8\%  & \textbf{82.0\%}
               & 0.101   & 0.071   & 0.161   & \textbf{0.066} 
               & 0.170   & 0.127   & 0.353   & \textbf{0.120}\\
                 
           5   & 73.3\%  & 80.6\%  & 59.7\%  & \textbf{81.9\%}
               & 0.103   & 0.071   & 0.164   & \textbf{0.066} 
               & 0.176   & 0.128   & 0.359   & \textbf{0.120}\\

        \midrule
            mean & 75.9\%  & 81.0\% & 58.6\%  & \textbf{82.4\%}
                 & 0.094   & 0.070  & 0.124    & \textbf{0.064} 
                 & 0.142   & 0.112  & 0.254    & \textbf{0.105}\\
        \bottomrule
        \end{tabular}
        \caption{ Quantitative comparisons on the task of \textbf{4D Shape Completion} from time-unevenly sampled point cloud sequences. The metrics of IoU, Chamfer distance, and Correspondence distance for all 6 time steps for the \emph{test} set (S1) \textbf{unseen motions (but seen individuals)} are reported.}
        \label{Tab:completion:motion}
        \end{center}
        \vspace{-5pt}
\end{table*}

\setlength{\tabcolsep}{3pt}
\begin{table*}[t]
    \vspace{-5pt}
	\renewcommand\arraystretch{1.2}
	\begin{center}
		\begin{tabular}{*{14}{c}}
			\toprule
			\multirow{2}*{Time step} & \multicolumn{4}{c}{IoU}   & \multicolumn{4}{c}{Chamfer} & \multicolumn{4}{c}{Correspond.}\\
			\cmidrule(lr){2-5} \cmidrule(lr){6-9} \cmidrule(lr){10-13}
			& OFlow &  C1 &  C2  & Ours  & OFlow &  C1 &  C2  & Ours  & OFlow &  C1 &  C2  & Ours \\
			\midrule
			\midrule
           0   & 71.5\%  & 72.5\% & 71.3\%  & \textbf{74.2\%}
               & 0.091   & 0.083  & 0.088   & \textbf{0.078} 
               & 0.085   & 0.078  & 0.087  & \textbf{0.075}\\
                 
           1   & 68.4\%   & 71.9\% & 63.0\%  & \textbf{73.4\%}
               & 0.105    & 0.086   & 0.126   & \textbf{0.080} 
               & 0.152    & 0.130   & 0.248   & \textbf{0.125}\\
                   
           2   & 67.5\%   & 71.9\% & 57.9\%  & \textbf{73.2\%}
               & 0.111    & 0.087   & 0.152   & \textbf{0.082} 
               & 0.177    & 0.140   & 0.325   & \textbf{0.137}\\
                   
           3   & 66.0\%    & 71.6\% & 54.1\%  & \textbf{72.7\%}
               & 0.117     & 0.088  & 0.173   & \textbf{0.083} 
               & 0.202     & 0.146  & 0.380   & \textbf{0.142}\\
                   
           4   & 64.4\%   & 71.1\% & 47.7\%  & \textbf{72.0\%}
               & 0.126    & 0.089  & 0.218   & \textbf{0.086} 
               & 0.236    & 0.167  & 0.500   & \textbf{0.162}\\
           
           5   & 64.0\%  & 71.1\% & 47.7\%  & \textbf{72.0\%}
               & 0.130   & 0.090   & 0.226   & \textbf{0.087} 
               & 0.246   & 0.171  & 0.521   & \textbf{0.164}\\
           
        \midrule
            mean & 67.0\%  & 71.7 \%   & 56.8\%  & \textbf{72.9\%}
                & 0.113    & 0.087    & 0.164    & \textbf{0.082} 
                & 0.183    & 0.139   & 0.344    & \textbf{0.134}\\
        \bottomrule
        \end{tabular}
        \caption{ Quantitative comparisons on the task of \textbf{4D Shape Completion} from time-unevenly sampled point cloud sequences. The metrics of IoU, Chamfer distance, and Correspondence distance for all 6 time steps for the \emph{test} set (S2) \textbf{unseen individuals} are reported.}
        \label{Tab:completion:individual}
        \end{center}
        \vspace{-5pt}
\end{table*}

    \subsection{Other Applications} 
    \label{SuppleOtherApp}
    In this section, we will investigate our pipeline for other applications, such as \emph{shape interpolation} and \emph{motion transfer}.
    For the shape interpolation, we find continuous transformations between the starting and ending mesh by firstly interpolating their latent embeddings $\mathbf{\hat{z}} $ = (1-w) * $\mathbf{z}_{start}$ + w* $\mathbf{z}_{end}, w \in [0, 1]$, and then predicting the vertex displacements. As shown in Fig.~\ref{fig:interpolation}, our method can accurately capture non-linear human motions.
    Concerning the motion transfer, we apply the predicted vertex displacements of human motions to the shape mesh.

}

{\small
\bibliographystyle{ieee_fullname}
\bibliography{egbib}
}

\end{document}